%% file: main.tex
\documentclass[10pt,twocolumn,letterpaper]{article}

\usepackage[pagenumbers]{cvpr} 

\usepackage{xcolor}

\input{preamble}

\definecolor{cvprblue}{rgb}{0.21,0.49,0.74}
\usepackage[pagebackref,breaklinks,colorlinks,allcolors=cvprblue]{hyperref}

\def\paperID{*****} 
\def\confName{CVPR}
\def\confYear{2026}

\title{UPLiFT: Efficient Pixel-Dense Feature Upsampling with Local Attenders}

\newcommand*{\affaddr}[1]{#1}
\newcommand*{\affmark}[1][*]{\textsuperscript{#1}}

\author{
Matthew Walmer\affmark[1] \quad
Saksham Suri\affmark[2] \quad
Anirud Aggarwal\affmark[1] \quad
Abhinav Shrivastava\affmark[1]\\
\affaddr{\affmark[1]University of Maryland, College Park} \quad
\affaddr{\affmark[2]Meta}\\
}

\begin{document}
\maketitle

\blfootnote{Code: \url{https://github.com/mwalmer-umd/UPLiFT/}}
\blfootnote{Web: \url{https://www.cs.umd.edu/~mwalmer/uplift/}}

\input{sec/0_abstract}
\input{sec/1_intro}
\input{sec/2_related}
\input{sec/3_arch}
\input{sec/4_experiments}
\input{sec/5_conclusion}

{
    \small
    \bibliographystyle{ieeenat_fullname}
    \bibliography{main}
}

\clearpage
\appendix
\input{sec_supp/A_predictive}
\input{sec_supp/B_generative}
\input{sec_supp/C_uplift_ablations}
\input{sec_supp/D_results}

\end{document}

%% file: preamble.tex
\newcommand{\myparagraph}[1]{\medskip\noindent\textbf{#1}}

\usepackage{booktabs}
\usepackage{graphicx}
\usepackage{caption}

\usepackage{amsmath}
\usepackage{algorithm}
\usepackage[noend]{algpseudocode}

\makeatletter
\def\BState{\State\hskip-\ALG@thistlm}
\makeatother

\newcommand\blfootnote[1]{%
  \begingroup
  \renewcommand\thefootnote{}\footnote{#1}%
  \addtocounter{footnote}{-1}%
  \addtocounter{Hfootnote}{-1}%
  \endgroup
}

\usepackage{url}
\urldef{\urlA}\url{https://huggingface.co/timm/vit_small_patch14_dinov2.lvd142m}

%% file: sec/0_abstract.tex
\vspace{-0.12in}
\begin{abstract}
The space of task-agnostic feature upsampling has emerged as a promising area of research to efficiently create denser features from pre-trained visual backbones. These methods act as a shortcut to achieve dense features for a fraction of the cost by learning to map low-resolution features to high-resolution versions.
While early works in this space used iterative upsampling approaches, more recent works have switched to cross-attention-based methods, which risk falling into the same efficiency scaling problems of the backbones they are upsampling.
In this work, we demonstrate that iterative upsampling methods can still compete with cross-attention-based methods; moreover, they can achieve state-of-the-art performance with lower inference costs. We propose \textbf{UPLiFT}, an architecture for \textbf{U}niversal \textbf{P}ixel-dense \textbf{Li}ghtweight \textbf{F}eature \textbf{T}ransforms. We also propose an efficient \textbf{Local Attender} operator to  overcome the limitations of prior iterative feature upsampling methods. This operator uses an alternative attentional pooling formulation defined fully locally. We show that our Local Attender allows UPLiFT to maintain stable features throughout upsampling, enabling state-of-the-art performance with lower inference costs than existing pixel-dense feature upsamplers. In addition, we apply UPLiFT to generative downstream tasks and show that it achieves competitive performance with state-of-the-art Coupled Flow Matching models for VAE feature upsampling.
Altogether, UPLiFT offers a versatile and efficient approach to creating denser features.
\end{abstract}

%% file: sec/1_intro.tex
\section{Introduction}
\label{sec:intro}

\input{figures/teaser_fig}

With the increasing spread of deep learning networks in various real-world applications, efficiency for visual information processing continues to be of key importance.
General-purpose, pre-trained visual backbones, such as DINO \cite{caron2021emerging, oquab2023dinov2, simeoni2025dinov3}, can be a driving force for efficiency, as these models provide a strong starting point for various visual applications, and reduce the total training cost needed to develop systems for new tasks. However, DINO and other Vision Transformer (ViT) \cite{dosovitskiy2020image} backbones face a fundamental problem: they must down-sample the spatial dimension to make visual tokens, which limits the final feature map density. For many applications, denser features are a necessity for effective performance, and while it is possible to increase the token density of ViTs to yield denser features \cite{amir2021deep}, the cost of self-attention operations scale quadratically with the number of visual tokens in the image. Thus, we face a fundamental trade-off between the benefits of denser features and the computational cost they incur.

For this reason, feature super-resolution methods have recently grown in popularity \cite{fu2024featup, suri2024lift, couairon2025jafar, huang2025loftup, wimmer2025anyup}. These methods act as task-agnostic add-ons to existing visual feature extracting backbones to directly transform their coarse, low-resolution features into high-resolution ones that preserve the original semantics. Such approaches have the potential to combine the benefits of dense features with the strength of pre-trained visual backbones, while avoiding the quadratic scaling cost of self-attention. Early approaches \cite{suri2024lift, fu2024featup} used multiple steps of iterative $2\times$ feature upsampling to push the coarse backbone features to pixel-scale-density. More recent approaches have switched to using cross-attention-based feature pooling to directly upsample to any desired output size. While these approaches offer strength and flexibility, they risk falling back into the pitfall of quadratic time scaling.

In this work, we examine the recent advances in task-agnostic feature super-resolution methods, and propose a new, efficient upsampler architecture called \textbf{UPLiFT} (\textbf{U}niversal \textbf{P}ixel-dense \textbf{Li}ghtweight \textbf{F}eature \textbf{T}ransforms). We draw inspiration from LiFT \cite{suri2024lift}, and we demonstrate that iterative upsampling methods are still a strong competitor with more recent cross-attention-based approaches. Not only does our UPLiFT model achieve state-of-the-art performance on several dense prediction tasks, it does so with lower inference costs than comparable recent methods.
To achieve this, we propose a new visual operator based on the observations of \cite{walmer2023teaching}, which showed that some ViT attention heads consistently learn to compute offset local attentional operations. As such, we propose a new \textbf{Local Attender} operator, an alternative formulation for attention which eschews the now standardized Query-Key-Value formulation and instead defines all operations based on local relative positions. We show that this localized attentional pooling mechanism achieves the same benefits of consistent feature semantics while avoiding the scaling costs of cross-attention.
As shown in Figure \ref{fig:teaser}, our UPLiFT method maintains roughly linear time scaling with respect to visual token count, while recent state-of-the-art cross-attention-based methods display quadratic time and high memory use.

\input{figures/tasks_figure}

Finally, we apply our UPLiFT model to a range of dense downstream tasks, including both predictive and generative tasks. In addition to pre-trained ViT features, we apply UPLiFT to upsample the latent features of Variational Autoencoders (VAE) \cite{kingma2013auto}, enabling applications in image super-resolution and efficient image generation. We show that in this domain, UPLiFT achieves competitive performance with state-of-the-art Coupled Flow-Matching (CFM) models \cite{schusterbauer2024fmboost}, while running with a much lower inference cost.
Overall, our contributions include:
\begin{itemize}
    \item \textbf{UPLiFT}, an efficient latent feature upsampler that can be applied to create high-quality, pixel-dense features.
    \item A \textbf{Local Attender} operator that reformulates attentional pooling into an efficient, locally-defined operation. We show that our Local Attender enables UPLiFT to maintaining consistent feature semantics, while avoiding the higher costs of comparable recent approaches.
    \item State-of-the-art performance on several dense predictive tasks, surpassing prior feature upsamplers while having faster inference speeds.
    \item An extended study of feature upsampling in the realm of generative tasks, showing the strength and efficiency of UPLiFT in comparison to a more computationally expensive Coupled Flow Matching method.
\end{itemize}

%% file: figures/teaser_fig.tex
\begin{figure}[t]
    \centering
    \includegraphics[width=1.0\linewidth]{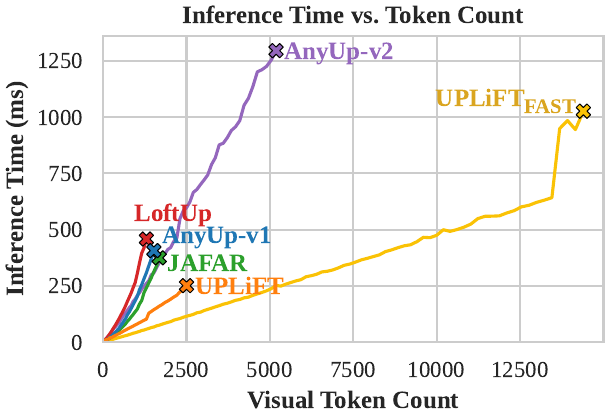}
    \includegraphics[width=1.0\linewidth]{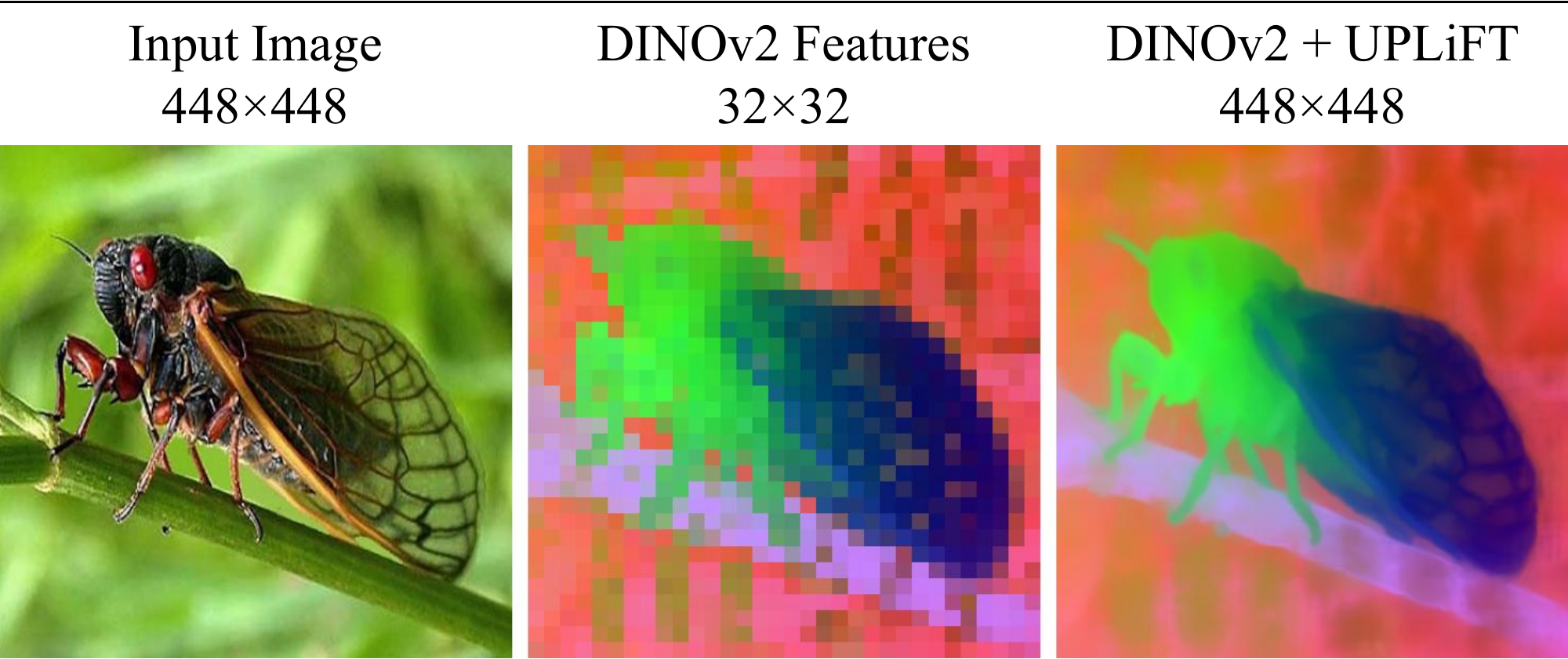}
    \vspace{-0.26in}
    \caption{\textbf{UPLiFT time-scaling and dense features.} We present \textbf{UPLiFT}, an efficient feature-upsampler that leverages our new \textbf{Local Attender} to extract semantically-stable, pixel-dense features. (Top) UPLiFT's inference time and memory scales linearly with the number of visual tokens, while most recent SOTA methods face quadratic scaling. (Bottom) PCA visualization of low-resolution DINOv2 features and pixel-dense UPLiFT features.}
    \label{fig:teaser}
    \vspace{-0.15in}
\end{figure}

%% file: figures/tasks_figure.tex
\begin{figure}[t]
    \centering
    \includegraphics[width=1.0\linewidth]{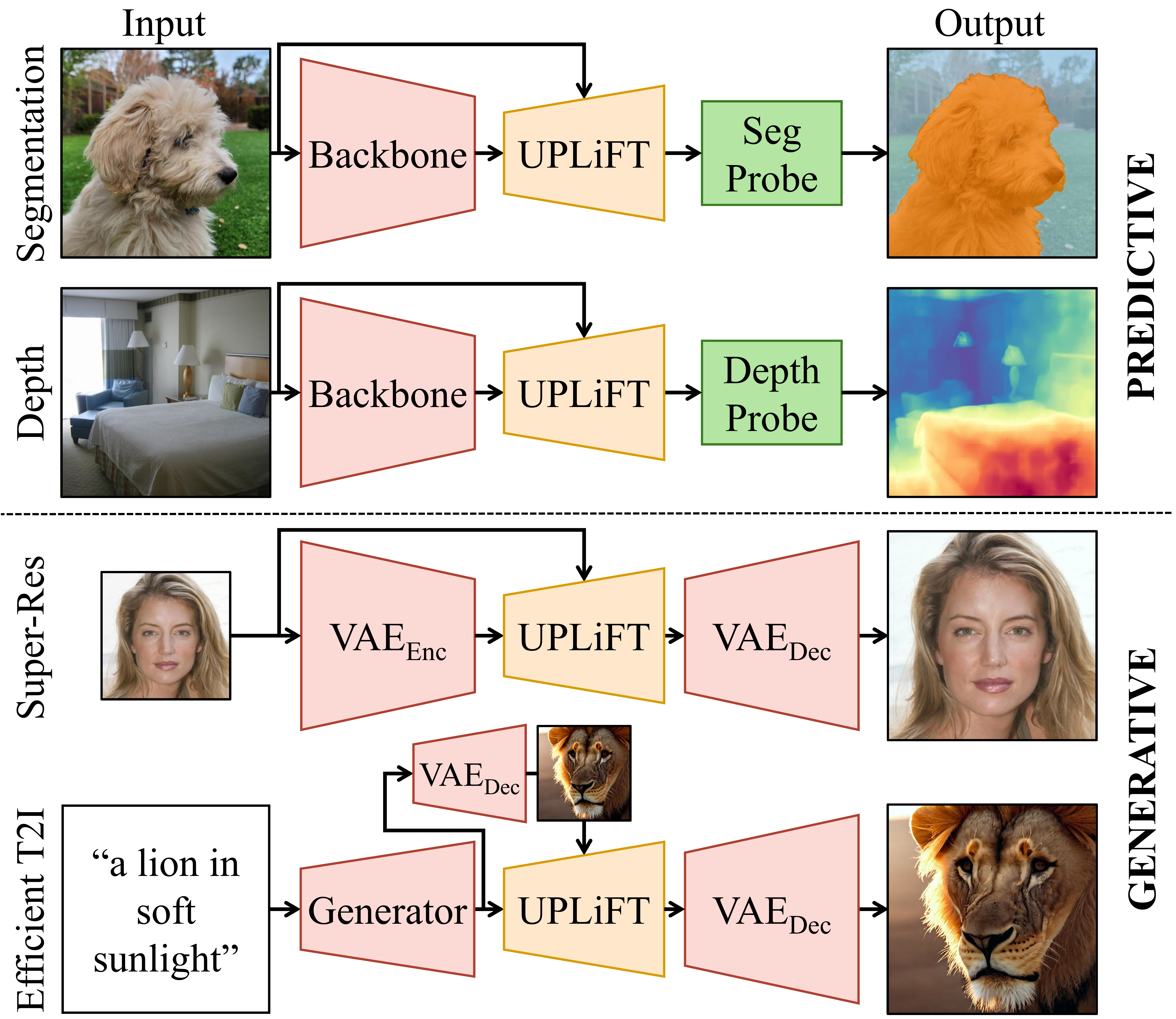}
    \vspace{-0.26in}
    \caption{\textbf{UPLiFT Tasks.} We demonstrate our UPLiFT feature upsampler for applications in both predictive and generative tasks. This includes semantic segmentation, monocular depth estimation, image super-resolution, and efficient text-to-image generation.}
    \label{fig:tasks}
    \vspace{-0.15in}
\end{figure}

%% file: sec/2_related.tex
\section{Related Works}
\label{sec:related}

\subsection{Feature Upsampling for Visual Backbones}

Higher-resolution visual features can be beneficial for many tasks, but extracting denser features comes with a larger computational cost. Feature-upsampling methods are a shortcut to create denser features from pre-trained, general-purpose visual backbones like DINO \cite{caron2021emerging, oquab2023dinov2, simeoni2025dinov3}, MoCo \cite{chen2020improved, chen2021empirical, he2020momentum}, MAE \cite{he2022masked}, CLIP \cite{radford2021learning}, and SigLip \cite{zhai2023sigmoid, tschannen2025siglip}.
In the past, many works have been proposed to design learnable feature adapters which are trained with supervision from a particular downstream task \cite{wang2019carafe, lu2022sapa, liu2023learning, zhou2024refreshed}. More recently, there has been an increasing number of methods proposed to output denser features in a task-agnostic way \cite{fu2024featup, suri2024lift, couairon2025jafar, huang2025loftup, wimmer2025anyup}. These methods are typically self-supervised, and are usually trained to upsample the features of a particular pre-trained backbone.
Such upsamplers can streamline development for dense tasks, by providing semantically rich and dense features directly out of the box.

Early works in this area \cite{fu2024featup, suri2024lift} used iterative upsampling with simple modules to create denser features.
FeatUp \cite{fu2024featup} proposed two feature upsampling methods, the first using an improved Joint Bilinear Upsampler (JBU) \cite{kopf2007joint} and the second using per-image Implicit Networks \cite{chen2021learning}.
The first method provides high-speed upsampling but lower upsampling quality, while the latter provides incredibly sharp and high-quality upsampled features, but has a high inference cost as it requires training an implicit model for each image.
LiFT \cite{suri2024lift} proposed a Lightweight Feature Transforming network with a simple self-supervised learning process to create $2\times$ upsampled features, with the potential for larger upsampling through iterative application of the same module. While this approach is simple and efficient, it has been shown that LiFT's iterative upsampling can lead to semantic drift and degraded features \cite{couairon2025jafar}. In this work, we propose a new iterative feature upsampler, UPLiFT, which overcomes the limitations of prior similar approaches.

\input{figures/arch_fig}

More recent works \cite{couairon2025jafar, huang2025loftup, wimmer2025anyup} have shifted to using cross-attention-based feature resampling. LoftUp \cite{huang2025loftup} and JAFAR \cite{couairon2025jafar} both propose a query-key-value (QKV) cross-attention \cite{vaswani2017attention} approach that combines high-resolution queries derived from an input image, with low-resolution keys and values derived from backbone features to perform flexible and direct upsampling. Moreover, this approach acts as an effective regularizer, ensuring the output features maintain a similar distribution to the input features. AnyUp \cite{wimmer2025anyup} further adapts the JAFAR architecture \cite{couairon2025jafar} into a backbone-agnostic model, with a special adapter layer that allows it to generalize to other feature extractors at inference time. These approaches show strong results, but their use of QKV cross-attention makes them less efficient, giving them quadratic time-scaling with respect to the number of visual tokens, as shown in Figure \ref{fig:teaser}. In contrast, our proposed UPLiFT architecture enjoys the same benefits of stable features while also remaining linear in time-scaling thanks to our Local Attender module.

AnyUp-v2 improves on the efficiency of the original AnyUp-v1 by using Neighborhood Attention (NATTEN) \cite{hassani2023neighborhood}, which also achieves linear scaling. We note from Figure \ref{fig:teaser} that our UPLiFT still has better scaling overall. Unlike NATTEN, our Local Attender does not use QKV attention, but instead uses a high-resolution guide feature map to derive the attention maps. In addition, we formulate all operations in terms of fixed directional offsets, removing the need for spatial embeddings.
We also recognize that UPLiFT's use of fixed-step upsamplers makes it slightly less flexible than cross-attention-based methods; however, given that most approaches simply aim to upsample to pixel-dense features, we believe this is not a significant detriment.

\subsection{Feature Upsampling for Generative Tasks}

The majority of prior works in task-agnostic feature upsampling have focused on visual feature extractors typically designed for predictive downstream tasks. Meanwhile, relatively little attention has been given to designing upsamplers for features for generative tasks. For example, the feature-spaces of Variational Autoencoders \cite{kingma2013auto} are now commonly used in Latent Diffusion models \cite{rombach2022high} to provided a compressed space for image generation. 
Because VAEs have decoders, upsampled VAE features can easily be decoded to enable tasks like image super-resolution and efficient text-to-image generation.
We note that both of these tasks are areas of significant research in their own right. For example, many methods exist that train primarily end-to-end generative architectures for super-resolution \cite{saharia2022image, chen2018fsrnet, dahl2017pixel, ledig2017photo, menon2020pulse}. 
However, for this work, we specifically wish to study methods which upsample VAE latent features, with image-super-resolution being a possible application of said features.
Meanwhile, extensive research has also been devoted to adapting pre-trained visual diffusion models to generate images at resolutions larger than their original training scale \cite{hwang2024upsample, qiu2025freescale, yang2025rectifiedhr, tragakis2024one, lin2025accdiffusion, li2024asgdiffusion}.
These methods may achieve a similar effect: adapting a pre-trained generator to create larger images, but they involve modifying the generator module/process itself, not the generated latents. These approaches also typically come with significantly increased generation costs, which feature upsampling could avoid.
Moreover, recent generator architectures like \cite{labs2025flux1kontextflowmatching} are designed to flexibly output features at a range of latent code sizes, but they still face the trade-off that visual feature extractors face: larger features come with a larger extraction/generation cost.

For this work, we aim to compare with methods that specifically learn to upsample VAE latent features. The state-of-the-art method in said space is Coupled Flow-Matching (CFM) \cite{schusterbauer2024fmboost}, which uses a powerful Flow-Matching model \cite{albergo2023stochastic, lipman2022flow, liu2022flow} conditioned on low-resolution VAE features to predict high-resolution VAE features. CFM is trained using a learning objective very similar to LiFT and UPLiFT, and it can be applied to upsample VAE latent codes either from encoded real images or from a generator like Stable Diffusion \cite{rombach2021highresolution}.
Unlike UPLiFT, CFM performs upscaling by first decoding the latent representation into pixel space, applying bilinear upscaling therein, and then re-encoding the result into latent space. The CFM module then refines the high-resolution features; however, the overall effect is still to produce high resolution latents from low resolution ones.
In this work, we show that UPLiFT can also learn to effectively upsample VAE features that produce output images with similar visual fidelity to CFM. Moreover, UPLiFT achieves this with a fraction of the parameters, training data, and inference cost of CFM.

%% file: figures/arch_fig.tex
\begin{figure*}[t]
    \centering
    \includegraphics[width=\linewidth]{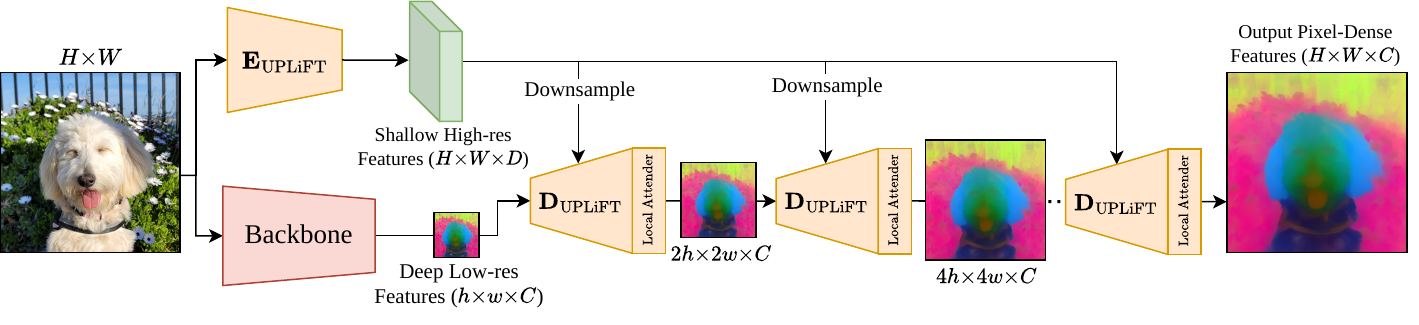}
    \vspace{-0.28in}
    \caption{\textbf{UPLiFT Inference.} At inference time, our UPLiFT Encoder ($\mathbf{E}_{\text{UPLiFT}}$) produces shallow but dense features to guide all subsequent upsampling steps. Iterative application of the UPLiFT Decoder ($\mathbf{D}_{\text{UPLiFT}}$) upsamples the low-resolution backbone features to pixel-density. Our proposed Local Attender module is integrated with the UPLiFT Decoder to maintain iterative feature consistency.}
    \label{fig:arch}
    \vspace{-0.1in}
\end{figure*}

%% file: sec/3_arch.tex
\section{UPLiFT Approach}
\label{sec:arch}

In this work, we propose an iteratively-growing feature upsampling approach inspired by LiFT \cite{suri2024lift}. While LiFT achieved effective $2\times$ feature upsampling, its approach to iterative upsampling for pixel-dense features could lead to semantic drift and degraded performance in downstream tasks.
In this work, we present UPLiFT, a new upsampler architecture that overcomes these issues to efficiently produce pixel-dense features while maintaining consistency in the feature distribution. Thanks to our Local Attender, UPLiFT's speed and memory requirements scale linearly with respect to the number of input visual patches/tokens, which leads to better speed and scaling properties than recent cross-attention-based methods.

\subsection{Architecture Overview}

We first illustrate our UPLiFT architecture in its inference-time configuration, as shown in Figure \ref{fig:arch}. UPLiFT has two primary modules: the UPLiFT encoder, denoted as $E_{\text{UPLiFT}}$, and the UPLiFT decoder, denoted as $D_{\text{UPLiFT}}$. $D_{\text{UPLiFT}}$ does the primary work of predicting upsampled features. Both are simple convolutional modules, which use strided convolution for downsampling or transpose convolution for upsampling. Note that we train a single compact $D_{\text{UPLiFT}}$ that performs $2\times$ upsampling, and that same module is applied multiple times to achieve pixel-dense upsampling. Meanwhile, this decoder is also guided by additional high-resolution features extracted by $E_{\text{UPLiFT}}$ from the input image. We follow the same intuition as LiFT: the input image already is a strong source of high-resolution information to guide feature upsampling. However, unlike LiFT, which required re-running its image encoder on every step with bilinearly upsampled input images, we instead design $E_{\text{UPLiFT}}$ such that it only runs once with the base input image. To do this, we make $E_{\text{UPLiFT}}$ into a shallow but dense encoder which preserves the original pixel density in its output features. These dense encoder features are downsampled with nearest-neighbors downsampling to the appropriate size to guide each upsampling step with $D_{\text{UPLiFT}}$. These design choices make UPLiFT both efficient and compact. However, as shown by \cite{couairon2025jafar}, this process of iterative upsampling risks introducing semantic drift, which we address in the following section with our Local Attender.

\subsection{Local Attender}

Preserving semantic consistency in the upsampled features is key to maintaining the strength of their representations. However, we wish to do this while avoiding quadratic scaling from QKV cross-attention. We propose a new formulation for a local attention operator, which achieves the same feature-regularization benefits of cross-attention while maintaining efficiency and linear-time-scaling. We draw inspiration from \cite{walmer2023teaching}, which showed that self-attention heads in ViTs often learn to attend to local positions with a fixed directional offset. Based on this, we propose \textbf{Local Attender}, which instead operates over a pre-set local neighborhood defined by a collection of fixed directional offsets. Under our Local Attender design, all attentional operations are defined relative to the current token's position, removing the need for positional embeddings. We illustrate our Local Attender operator in Figure \ref{fig:localatt}, and define it as follows.

First, we assume two input feature maps, a ``Guide Feature'' $G$ and a ``Value Feature'' $V$. The Local Attender operator uses $G$ to guide a linear resampling of $V$ to produce a new feature map output. Specifically, $G$ is used to predict an attention map over a local neighborhood around each token in $V$.
For an initial case, assume $G$ and $V$ have the same spatial dimensions, $H{\times}W$, but different channel depths, $C_G$ and $C_V$.
Next, we define a fixed neighborhood $\mathbf{N}$, which is a set of $2D$ directional integer offsets, $(i,j)$. For a given visual token $V_{x,y}$ in $V$, it can only attend to positions $V_{x+i,y+j}$ for $(i,j)\in\mathbf{N}$. In this way, the neighborhood $\mathbf{N}$ defines the range and size of the local attentional operations computed by the Local Attender. Moreover, the size and pattern of $\mathbf{N}$ can be flexibly defined. We present experiments with several neighborhood designs in the Appendix \ref{supp:ablations}. In general, let $||\mathbf{N}||=n$.

\input{figures/local_attender_arch}

Next, we apply a $1{\times}1$ convolutional layer to $G$ to convert it to shape $H{\times}W{\times} n$. This convolutional layer is the only learnable element of the Local Attender. We then apply position-wise softmax to this feature, creating the ``Attender Map'' $A$. For a given $(x,y)$ position in $A$, the value $A_{x,y,k}$ represents the attentional weight to apply to the feature at $V_{x+i_k,y+j_k}$ where $(i_k,j_k)$ is the $k^\text{th}$ offset in $\mathbf{N}$. In practical terms, we compute a set of offset feature maps by applying the offsets of $\mathbf{N}$ to Value Feature $V$ with replication padding. We then multiply these maps by the Attender Map $A$ and we sum along the newly created ``neighborhood-dimension'' to produce the final locally-attended features of shape $H{\times}W{\times}C_V$. 
Let $T$ equal the number of spatial tokens in $G$. Then the memory and compute cost of the Local Attender scales as $O(nT)$, or linearly in $T$ as $n$ is a constant.

Finally, we relax an initial assumption to allow Local Attender to act as an upsampling operator. Instead, assume $V$ has shape $H{\times}W{\times}C_V$ and guide $G$ has shape $cH{\times} cW{\times}C_G$, where $c\in\mathbb{Z}$. We first group the tokens of G into an $H{\times}W$ grid of cells, each cell being of shape $c{\times}c$. For all tokens in a given cell $C_{x,y}$ with $(x,y)\in H{\times}W$, the neighborhood they can attend to is defined around the corresponding Value token at $V_{x,y}$. As such, the size of the neighborhoods remains the same, but each token in $V$ now corresponds to a $c{\times}c$ cell in $G$. Once again, the output features are linear sums of features in $V$, ensuring feature consistency with respect to $V$. The output map size is determined by $G$, with the final output being $cH{\times}cW{\times}C_V$.

\input{figures/training_fig}

\input{tables/linear_probes_table}

We integrate our Local Attender operator as the final step of our UPLiFT decoder, with the initial decoder output as $G$ and the original backbone feature as $V$. We show that the Local Attender ensures the preservation of the backbone feature distribution, which allows UPLiFT to efficiently create semantically-strong, pixel-dense features. In the following section, we show that this leads to state-of-the-art performance with faster inference speeds than recent methods.

\subsection{UPLiFT Training}
\label{sec:archtrain}

We train UPLiFT with a low-to-high-res self-supervised feature prediction objective based on LiFT, though we adapt our training to explicitly include multiple upsampling steps during training. We illustrate our UPLiFT training process in Figure \ref{fig:training}. 
Let $B$ represent the frozen, pre-trained visual backbone. Let $E_{U}$ and $D_{U}$ denote the UPLiFT Encoder and Decoder, which are trained jointly in this process. Let $H{\times}W$ denote the maximum size an image is loaded at. We refer to this high-resolution image as $I$. We also specify a ``training depth'' $d$, which describes the scale of downsampling applied to the images that are input to UPLiFT. For a given depth $d$, the ground-truth image will be downsampled by a factor of $2^d$ to size $(H/2^d){\times} (W/2^d)$. We denote this low-resolution image as $I'$. Next, we pass both $I$ and $I'$ through $B$ to extract low-resolution and high-resolution features $F=B(I)$ and $F'=B(I')$. Additionally, $I'$ is passed through $E_{U}$ to produce features $F'_E$ which are used to guide upsampling. Note that with our dense UPLiFT encoder, $F'_E$ has the same spatial resolution as $I'$, while $F'$ has lower-resolution with size $(H/2^dp){\times} (W/2^dp)$, where $p$ is the patch size.

Next, we perform feature upsampling. To start, $D_{U}$ receives the low resolution feature $F'$ and guiding feature $F'_E$ which is down-sampled to $2\times$ the scale of $F'$. $D_{U}$ predicts upsampled features with resolution $(H/2^{d-1}p){\times}(W/2^{d-1}p)$, which are then provided to the Local Attender to upsample $F'$. The result is $F'_{2{\times}}$, which is two times larger than the original low-resolution features. This process is repeated $d$ times, until we produce $F'_{2^d{\times}}$ with size $(H/p){\times} (W/p)$, the same as $F$. We then compute the reconstruction loss as the $L2$ distance between the upsampled feature and the high-resolution feature:
\begin{equation}
\label{eq:simple}
\mathcal{L}_\text{simple} = \mathcal{D}_{L2}(F'_{2^d{\times}},F)
\end{equation}
This simple learning objective explicitly incorporates multiple applications of $D_U$ during training, which helps UPLiFT learn to create pixel-dense features at inference time. However, this training does not necessarily need to be applied with $d = \log_2(p)$, as shallower depths can be sufficient to train UPLiFT.
In addition, for each $D_{U}$ upsampling step, an additional intermediate feature map can be extracted to derive an additional loss term. Denote the full set of upsampled feature maps as: 
\begin{equation}
\label{eq:predfeats}
\begin{split}
\mathbf{F'} & = \{F'_{2{\times}}, F'_{4{\times}},~...~,F'_{2^{d-1}{\times}},F'_{2^d{\times}}\} \\
            & =\{F'_{1/2^{d-1}},F'_{1/2^{d-2}},~...~,F'_{1/2},F'_{1/1}\}
\end{split}
\end{equation}
We also compute intermediate ground truth feature maps by downsampling $I$ to intermediate resolutions, with $F_{1/2^{k}} = B(I_{1/2^k})$, to give the set of target feature maps:
\begin{equation}
\label{eq:gtfeats}
\mathbf{F}=\{F_{1/2^{d-1}},F_{1/2^{d-2}},~...~,F_{1/2},F_{1/1}\}
\end{equation}
Then we can compute the total loss for depth $d$ with all intermediate feature maps as:
\begin{equation}
\label{eq:depthloss}
\mathcal{L}_{d} = \sum_{k=1}^{d} \mathcal{D}_\text{L2}(F'_{1/2^{d-k}},F_{1/2^{d-k}})
\end{equation}
In practice, we find UPLiFT achieves the best performance when multiple depths are used during training. For our primary results, we use $d\in \mathbf{D}:=\{1,2,3\}$ for each training step. We present ablations with additional configurations in Appendix \ref{supp:ablations}.
Overall, our final UPLiFT learning objective can be denoted as:
\begin{equation}
\label{eq:upliftloss}
\mathcal{L}_\text{UPLiFT} = \sum_{d\in \mathbf{D}} \mathcal{L}_{d}
\end{equation}

\subsection{Accelerating UPLiFT Inference}
\label{sec:upliftfast}

We apply several inference-time optimization strategies to further accelerate UPLiFT, denoted ``UPLIFT$_{\text{FAST}}$'' in Section \ref{sec:exppred}. These improvements do not change the underlying model architecture or training, and can be applied to any pretrained UPLiFT model. These optimizations improve speed and reduce memory usage without impacting downstream task performance. Additional details can be found in Appendix \ref{supp:predictive}.

\input{figures/cfm_comparison_fig}

\input{tables/coco_laion_table}
\input{tables/super_res_table}

%% file: figures/local_attender_arch.tex
\begin{figure}[t]
    \centering
    \includegraphics[width=1.0\linewidth, trim=0 0 0.5cm 0, clip=true]{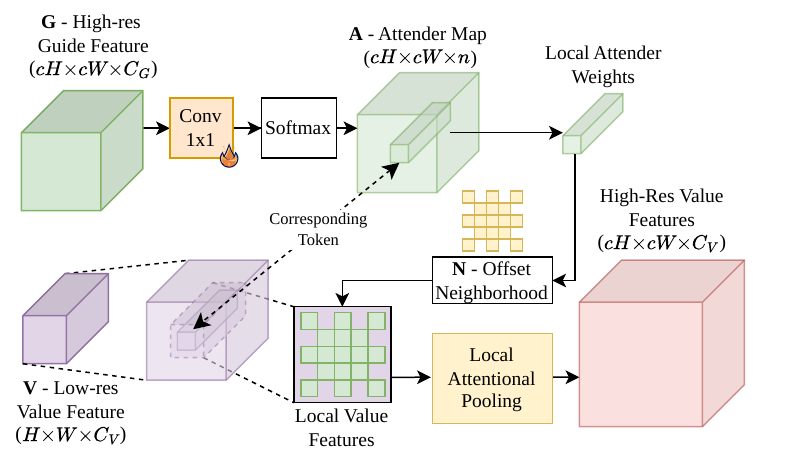}
    \vspace{-0.3in}
    \caption{\textbf{Local Attender Operator.} We propose a streamlined and efficient local attention operator, which gathers features over a set neighborhood defined by fixed direction offsets.}
    \label{fig:localatt}
    \vspace{-0.15in}
\end{figure}

%% file: figures/training_fig.tex
\begin{figure}[t]
    \centering
    \includegraphics[width=1.0\linewidth, trim=1.5cm 0 0 0, clip=true]{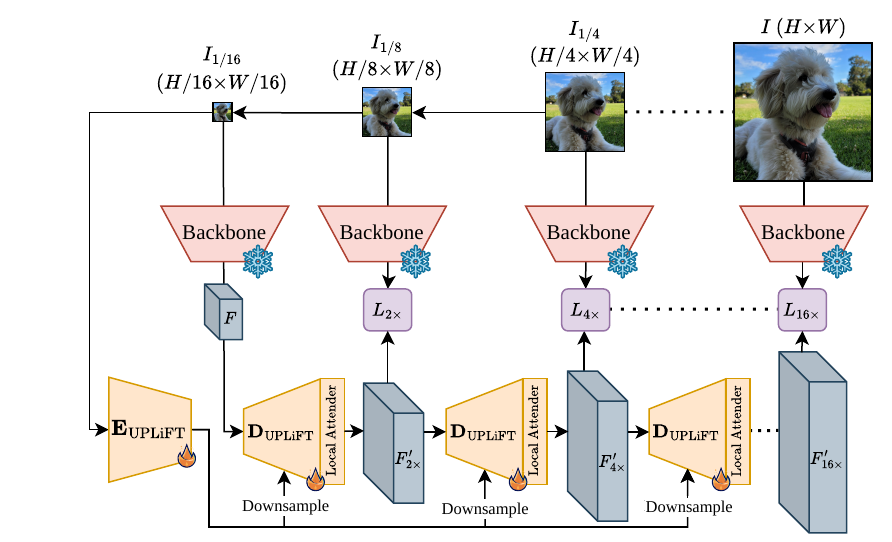}
    \vspace{-0.26in}
    \caption{\textbf{UPLiFT Training.} UPLiFT uses a multi-step training strategy where the feature reconstruction loss is applied at all intermediate steps. All lift decoders ($\mathbf{D}_{\text{UPLiFT}}$) share the same weights.}
    \label{fig:training}
    \vspace{-0.15in}
\end{figure}

%% file: tables/linear_probes_table.tex
\begin{table*}[t]
  \caption{\textbf{Segmentation and Depth Estimation.} UPLiFT surpassed all prior upsampling methods in semantic segmentation on four datasets. For depth estimation, it is second best for $\delta_1$ and tied for best in RMSE. UPLiFT achieves these scores with faster inference speeds than recent SOTA methods. Best results are marked in \textbf{bold} and second best are \underline{underlined}.}
  \vspace{-0.12in}
  \label{tab:lp}
  \centering
  \resizebox{1.0\linewidth}{!}{
    \begin{tabular}{@{}lcc|cc|cc|cc|cc|cc@{}}
      \toprule
      \multicolumn{3}{@{}c}{} &
      \multicolumn{8}{c@{}}{Semantic Segmentation} &
      \multicolumn{2}{c@{}}{Depth Estimation} \\
      \cmidrule(r{2pt}){4-11}\cmidrule(l{2pt}){12-13}
      \multicolumn{3}{c}{Upsampler} &
      \multicolumn{2}{c}{COCO} &
      \multicolumn{2}{c}{VOC} &
      \multicolumn{2}{c}{ADE20K} &
      \multicolumn{2}{c}{Cityscapes} &
      \multicolumn{2}{c}{COCO} \\
      \cmidrule(r{2pt}){1-3}\cmidrule(l{2pt}r{2pt}){4-5}\cmidrule(l{2pt}r{2pt}){6-7}\cmidrule(l{2pt}r{2pt}){8-9}\cmidrule(l{2pt}r{2pt}){10-11}\cmidrule(l{2pt}){12-13}
      Method & Params\,(M) & Time\,(ms) &
      mIoU $\uparrow$ & Acc $\uparrow$ & mIoU $\uparrow$ & Acc $\uparrow$ &
      mIoU $\uparrow$ & Acc $\uparrow$ & mIoU $\uparrow$ & Acc $\uparrow$ &
      $\delta_1 \uparrow$ & RMSE $\downarrow$ \\
      \midrule
      Nearest           & -- & 0.6 & 56.41 & 77.11 & 78.29 & 94.37 & 37.86 & 72.17 & 54.87 & 90.56 & 56.66 & 0.73 \\
      Bilinear          & -- & 2.8 & 59.41 & 79.28 & 81.62 & 95.44 & 40.43 & 74.12 & 59.71 & 92.56 & 58.83 & 0.68 \\
      \midrule
      LiFT-$2\times$   & 1.2 & 3.8 & 58.28 & 78.97 & 82.46 & 95.73 & 39.74 & 73.79 & 61.07 & 93.09 & 57.17 & 0.70 \\
      LiFT  & 1.2 & 51.9 & 57.42 & 78.46 & 80.97 & 95.37 & 38.95 & 73.34 & 61.98 & 93.61 & 55.07 & 0.73 \\
      FeatUp            & 0.2 & 109.6 & 61.77 & 80.99 & 83.52 & 96.06 & 42.07 & 75.52 & 60.50 & 93.12 & 60.01 & 0.66 \\
      LoftUp            & 4.3 & 223.5 & \underline{62.19} & \underline{81.35} & 84.63 & 96.33 & 42.16 & 75.79 & 62.09 & 93.55 & 58.69 & 0.68 \\
      JAFAR             & 0.7 & 111.7 & 61.71 & 81.01 & 84.38 & 96.22 & 41.96 & 75.43 & 61.89 & 93.52 & 60.59 & 0.65 \\
      AnyUp-v1             & 0.9 & 146.7 & 62.08 & 81.31 & 84.33 & 96.23 & 42.25 & 75.80 & 61.33 & 93.44 & \textbf{61.32} & \textbf{0.63} \\
      AnyUp-v2          & 0.8 & 153.1 & 61.95 & 81.24 & 84.09 & 96.18 & 42.13 & 75.71 & 60.68 & 93.27 & \underline{61.26} & \underline{0.64} \\
      \midrule
      \textbf{UPLiFT}            & 0.8 & 79.4 & \textbf{62.55} & \textbf{81.57} & \textbf{85.21} & \textbf{96.51} & \underline{42.97} & \underline{76.00} & \textbf{65.38} & \underline{94.41} & 61.16 & \textbf{0.63} \\
      \textbf{UPLIFT$_{\text{FAST}}$} & 0.8 & 41.6 & \textbf{62.55} & \textbf{81.57} & \underline{85.08} & \underline{96.48} & \textbf{43.00} & \textbf{76.02} & \underline{65.36} & \textbf{94.42} & 61.20 & \textbf{0.63} \\
      \bottomrule
    \end{tabular}
  }
  \vspace{-0.15in}
\end{table*}

%% file: figures/cfm_comparison_fig.tex
\begin{figure*}[t]
    \centering
    \includegraphics[width=1.0\linewidth]{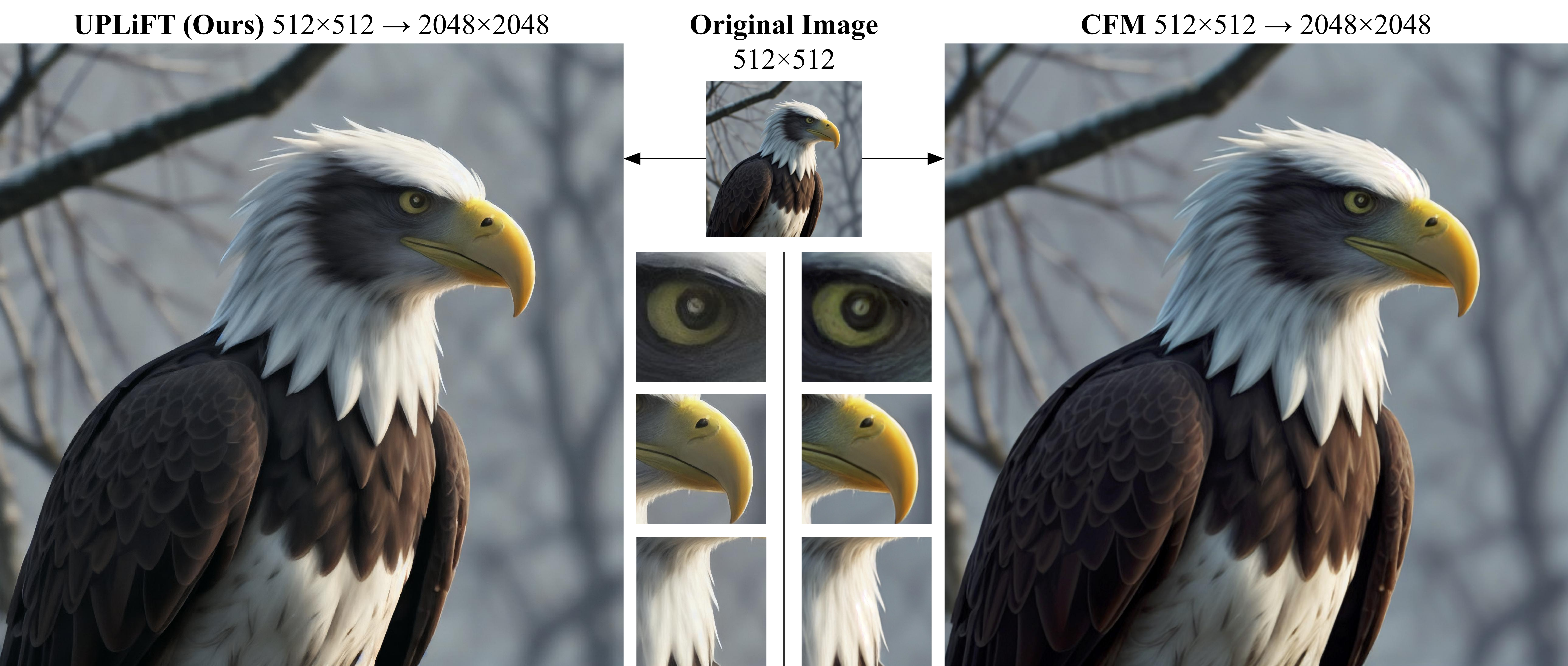}
    \vspace{-0.27in}
    \caption{\textbf{2048$\times$2048 Image Generation Comparison with CFM.} We show that UPLiFT achieves comparable visual upsampling quality to CFM \cite{fu2024featup}, while using only $1/6^{\text{th}}$ the network parameters, $1/200^{\text{th}}$ the training data, and only 2 iterative upsampling steps.}
    \label{fig:cfmcomparison}
    \vspace{-0.05in}
\end{figure*}

%% file: tables/coco_laion_table.tex
\begin{table*}[t!]
  \centering
  \begin{minipage}[t]{0.475\linewidth}
      \renewcommand{\tabcolsep}{3pt}
      \renewcommand{\arraystretch}{1.23}
      \caption{\textbf{Zero-shot COCO-5k 512$\to$1024.}
      UPLiFT achieves superior FID to CFM for COCO text-to-image upsampling, while running with significantly faster inference speeds. Like \cite{schusterbauer2024fmboost}, we measure speed for this task with batch size 4. CFM is run with 20 steps. 
      $^\dagger$Row adapted from \cite{schusterbauer2024fmboost}}
      \vspace{-0.12in}
      \label{tab:coco5k}
      \centering
      \resizebox{\linewidth}{!}{
        \begin{tabular}{@{}ccccccc@{}}
          \toprule
          \multicolumn{2}{@{}c}{Diffusion Model}  & \multicolumn{3}{c}{Upsampler}  & \multicolumn{2}{c@{}}{Performance} \\
          \cmidrule(r){1-2}\cmidrule(lr){3-5}\cmidrule(l){6-7}
          Model & Steps  & Method & Params (M) & Resolution  & s/img $\downarrow$ & FID $\downarrow$ \\
          \midrule
          SD1.5 & 25  & -- & -- & $512$  & 3.96 & 24.88 \\
          \midrule
          SD1.5 & 25  & CFM-20$^\dagger$ & 306 & $512 \to 1024$  & 8.79 & 28.81 \\
          SD1.5 & 25  & \textbf{UPLiFT}  & 53  & $512 \to 1024$  & \textbf{5.15} & \textbf{24.23} \\
          \midrule
          SDXL & 50  & -- & -- & $1024$  & 39.84 & 24.07 \\
          \bottomrule
        \end{tabular}
      }
    \end{minipage}
  \hfill
  \begin{minipage}[t]{0.475\linewidth}
  \renewcommand{\arraystretch}{0.96}
  \renewcommand{\tabcolsep}{3.5pt}
  \caption{\textbf{Zero-shot reLAION-400M-5k 512$\to$1024.}
  UPLiFT demonstrates strong upsampling for text-to-image, surpassing CFM in all three metrics, with faster inference. CFM is run with 4 or 40 steps.
  $^*$As CFM does not report diffusion steps, we include two options for comparison.
  Speed measured with batch size 1 as in \cite{schusterbauer2024fmboost}.
  $^\dagger$Rows adapted from \cite{schusterbauer2024fmboost}.}
  \vspace{-0.12in}
  \label{tab:laion5k}
  \centering
  \resizebox{\linewidth}{!}{
    \begin{tabular}{@{}lcccccccc@{}}
      \toprule
      \multicolumn{2}{@{}c}{Diffusion Model}  & \multicolumn{3}{c}{Upsampler}  & \multicolumn{4}{c@{}}{Performance} \\
      \cmidrule(r){1-2}\cmidrule(lr){3-5}\cmidrule(l){6-9}
      Model & Steps  & Method & Params (M) & Resolution  & s/img $\downarrow$ & CLIP $\uparrow$ & FID $\downarrow$ & p-FID $\downarrow$ \\
      \midrule
      SD1.5 & 25  & -- & -- & $512$  & 1.27 & 31.67 & 22.32 & -- \\
      SD1.5 & 40  & -- & -- & $512$  & 1.96 & 31.87 & 21.90 & -- \\
      \midrule
      SD1.5 & *  & CFM-4$^\dagger$  & 306 & $512{\to}1024$  & 2.72 & 26.16 & 21.61 & \underline{15.83} \\
      SD1.5 & *  & CFM-40$^\dagger$ & 306 & $512{\to}1024$  & 3.16 & 26.14 & 21.67 & 15.96 \\
      SD1.5 & 25  & \textbf{UPLiFT} & 53  & $512{\to}1024$  & \textbf{1.56} & \underline{31.00} & \underline{21.17} & 16.10 \\
      SD1.5 & 40  & \textbf{UPLiFT} & 53  & $512{\to}1024$  & \underline{2.27} & \textbf{31.17} & \textbf{20.73} & \textbf{15.49} \\
      \midrule
      LCM SDXL & 4  & -- & -- & $1024$  & 0.72 & 31.06 & 28.05 & 24.13 \\
      \bottomrule
    \end{tabular}
  }
\end{minipage}
\vspace{-0.2in}
\end{table*}

%% file: tables/super_res_table.tex
\begin{table*}[t]
  \caption{\textbf{Zero-shot Image Super Resolution 256$\to$1024 on 5k samples of FacesHQ and LHQ.}
  UPLiFT achieves superior scores in SSIM and PNSR for LHQ, and better SSIM for FacesHQ.
  We note that UPLiFT achieves these results with a single general-purpose module, while CFM uses dataset-specific trained modules with significantly more parameters and iterations.
  $^\dagger$Row adapted from \cite{schusterbauer2024fmboost}}
  \vspace{-0.12in}
  \label{tab:super-res}
  \centering
  \resizebox{1.0\textwidth}{!}{
    \begin{tabular}{@{}lc|ccccc|ccccc@{}}
      \toprule
      \multicolumn{2}{@{}c}{Upsampler} &
      \multicolumn{5}{c}{FacesHQ} &
      \multicolumn{5}{c@{}}{LHQ} \\
      \cmidrule(r{2pt}){1-2}\cmidrule(l{2pt}r{2pt}){3-7}\cmidrule(l{2pt}){8-12}
      Method & Iters & 
      Params (M) & SSIM $\uparrow$ & PSNR $\uparrow$ & FID $\downarrow$ & p-FID $\downarrow$ &
      Params (M) & SSIM $\uparrow$ & PSNR $\uparrow$ & FID $\downarrow$ & p-FID $\downarrow$ \\
      \midrule
      Bilinear & 1    &  --  & 0.72      & 21.74      & 108.12     & 176.90    & --   & 0.63      & 22.09      & 147.65     & 224.70 \\
      Nearest  & 1    &  --  & 0.66      & 20.48      & 193.08     & 332.28    & --   & 0.60      & 21.08      & 241.55     & 288.66 \\
      CFM$^\dagger$   & 50   & 113  & \underline{0.82} & \textbf{30.40}    &  \textbf{1.36}    & \textbf{1.62}    & 306 & \underline{0.69} & \underline{25.69} & \textbf{2.27}     & \textbf{2.38} \\
     \textbf{UPLiFT}  & 2    &  53  & \textbf{0.84}    & \underline{29.93} &  \underline{5.20} & \underline{6.62} & 53  & \textbf{0.73}    & \textbf{26.70}    &  \underline{5.29} & \underline{7.23} \\
      \bottomrule
    \end{tabular}
  }
\vspace{-0.12in}
\end{table*}

%% file: sec/4_experiments.tex
\section{UPLiFT for Predictive Tasks}
\label{sec:exppred}

\myparagraph{Experimental Methods.}
We start with experiments focused on dense predictive tasks: semantic segmentation and monocular depth estimation.
We compare with other task-agnostic feature-upsampling methods LiFT\,\cite{suri2024lift}, Featup\,\cite{fu2024featup}, LoftUp\,\cite{huang2025loftup}, JAFAR\,\cite{couairon2025jafar}, AnyUp-v1, and AnyUp-v2\,\cite{wimmer2025anyup}. We also compare with simple baselines of bilinear and nearest-neighbor upsampling. 
To compare with prior works, we select DINOv2-S/14\,\cite{oquab2023dinov2} as our primary standard backbone for experiments. For all tasks in this section, we use an UPLiFT model trained for one epoch on the ImageNet-1K dataset \cite{deng2009imagenet}, with a maximum ground truth image size of 448 and a maximum input image size of 224. We train using our multi-step loss (Section \ref{sec:archtrain}) with 3 depth levels. We present results with the same UPLiFT model running in both regular mode and fast mode (Section \ref{sec:upliftfast}).

We build on the experimental protocols of \cite{couairon2025jafar} and train linear probes models on top of the upsampled DINOv2-S/14 features produced by each method. In addition, following the example of \cite{wimmer2025anyup}, we make a small correction to the learning rate schedule of the official JAFAR evaluation suite. For this reason, we have recomputed all the baseline methods to provide a fair comparison. All methods are run in a pixel-dense upsampling mode. Following the example of \cite{couairon2025jafar}, we also run LiFT in a $2{\times}$ upsampling mode followed by bilinear upsampling to pixel-dense features, which we denote as LiFT-$2{\times}$. All evaluations use $448{\times}448$ input images and $448{\times}448$ upsampled features. For each method, we also report the number of parameters and the average upsampling time for a single image's feature as measured on one NVIDIA A5000 GPU.
Please see the Appendix \ref{supp:predictive} for additional details.

\myparagraph{Segmentation.}
We present semantic segmentation results in Table \ref{tab:lp} for four datasets: COCO-Stuff\,\cite{lin2014microsoft}, Pascal VOC\,\cite{everingham2015pascal}, ADE20k\,\cite{zhou2019semantic}, and Cityscapes\,\cite{cordts2016cityscapes}. First, our results confirm the observations of \cite{couairon2025jafar}, that iteratively running LiFT to produce pixel-dense features leads to degraded performance in this task. We find that a single LiFT upsampling iteration (LiFT-$2{\times}$) yields better performance than iterative upsampling to pixel-dense features. We also note that AnyUp-v2's performance closely matches AnyUp-v1, although it is slightly lower in some cases. This may be attributed to changes in the model's implementation or training method, which we discuss more in Appendix \ref{supp:predictive}. We also note that, at this inference resolution, AnyUp-v2 is not faster than AnyUp-v1, though we examine their scaling properties more later in this section.
Next, we see that UPLiFT achieves state-of-the-art performance, with the highest mIoU and Accuracy scores for all four datasets, even surpassing recent methods that use cross-attention feature pooling. This result demonstrates that our Local Attender approach is an effective method for feature pooling and upsampling. Furthermore, we see that the performance of UPLIFT$_{\text{FAST}}$ is nearly identical to UPLiFT, which shows that our inference-time acceleration methods do not impact downstream performance.
Finally, we see that UPLiFT has faster inference speed than all high-performing pixel-dense upsamplers, and that UPLIFT$_{\text{FAST}}$ almost doubles the inference speed of UPLiFT. While other baselines like Nearest, Bilinear, and LiFT-$2{\times}$ achieve faster speeds, they also have inferior performance, meaning UPLiFT provides an optimal combination of speed and feature quality.

\myparagraph{Depth Estimation.}
We next apply UPLiFT to monocular depth estimation on COCO-Stuff. Like \cite{couairon2025jafar}, we report the $\delta_1$ score, which is a thresholded accuracy score, and the Root Mean Square Error (RMSE). Predicting visual depth from monocular cues requires a broader understanding of the entire image. In theory, the cross-attention-based methods, LoftUp, JAFAR, and AnyUp, should have an advantage in this task, as their feature pooling approach can gather broader information from wider image regions. However, we find that UPLiFT ties with AnyUp-v1 for the lowest RMSE score, and is beaten only by AnyUp-v1 and v2 for $\delta_1$ score. These results demonstrate that our UPLiFT features, which are upsampled using only local information through our Local Attender, can still derive sufficient global information from the backbone features to achieve competitive performance in depth estimation.

\myparagraph{Efficiency and Scaling of UPLiFT.}
As shown by Table \ref{tab:lp}, UPLiFT achieves faster inference speeds than recent state-of-the-art methods using cross-attention pooling, and UPLIFT$_{\text{FAST}}$ further improves this difference. 
Said speeds are measured for an image size of $448{\times}448$ with a patch size of $14$, which yields $1024$ visual tokens. As recent methods have moved to using cross-attention for feature scaling, they now face quadratic time and memory scaling with the number of input tokens. Meanwhile, our UPLiFT method maintains linear scaling, leading to even faster performance for larger image sizes. This is essential, as producing high-resolution image features can easily become a memory-intensive process. To demonstrate this, we test UPLiFT, LoftUp, JAFAR, and AnyUp for gradually increasing image sizes, until they hit a memory limit of $24$ GB on an A5000 GPU. We assume a patch size of $16$ and run all methods in a pixel-dense upsampling configuration. We report the average inference speed against the number of visual tokens in Figure \ref{fig:teaser}. LoftUp, JAFAR, and AnyUp-v1 all show signs of quadratic scaling and run out of memory at ${\sim}1500$ tokens. AnyUp-v2 with NATTEN achieves linear scaling and reaches ${\sim}5000$ tokens.
Both UPLiFT and UPLIFT$_{\text{FAST}}$ have linear time scaling and are significantly faster than the other baselines. Moreover, the maximum token count that can be processed by UPLIFT$_{\text{FAST}}$ far surpasses all other methods, exceeding $14000$ tokens.
We note that the time jump observed for UPLIFT$_{\text{FAST}}$ at ${\sim}13500$ tokens is likely due to GPU memory allocation issues. Please see Appendix \ref{supp:results} for additional analysis.

\section{UPLiFT for Generative Tasks}
\label{sec:expgen}

\myparagraph{Experimental Methods.}
In this section, we extend UPLiFT to generative tasks by applying it to VAE latent features, enabling efficient image generation and super-resolution.
To preserve the feature distribution for generative tasks, we find that a larger UPLiFT model is a necessity, but our larger model still only has one half or one sixth the parameters of the comparable CFM models.
We train UPLiFT for 5 epochs on Unsplash-Lite \cite{unsplash_data}, which contains 25k high-quality images. For comparison, CFM's general-purpose model is trained on the Unsplash dataset with over 5 Million images.
Using a maximum ground-truth image size of 1024, we train across 4 depth levels using the multi-step iterative loss introduced in Section \ref{sec:archtrain}.
Both methods use the Stable Diffusion 1.5 \cite{rombach2021highresolution} VAE to perform encoding and decoding between pixel and latent space.

We follow the protocols of \cite{schusterbauer2024fmboost} across two generative tasks: text-to-image diffusion upscaling and image super-resolution.
We first use 5k random samples from COCO 2014 and reLAION-400M to evaluate UPLiFT's ability to upscale diffusion features. This task effectively increases Stable Diffusion 1.5's output size from $512{\times}512$ to $1024{\times}1024$.
We compare with CFM, as well as SDXL and LCM-LoRA SDXL, which natively generate at $1024{\times}1024$, to provide an additional point of reference for visual quality from high-resolution latents.
We also evaluate UPLiFT for image super-resolution on the FacesHQ and LHQ datasets, again following the protocols of \cite{schusterbauer2024fmboost}.
Additionally, we compare against two lightweight baselines: direct bilinear and nearest neighbors upsampling in latent space.
Reported latency is measured on a single NVIDIA A100 GPU.
Please see Appendix \ref{supp:generative} for additional details on the model, training, and tasks.

\myparagraph{Efficient High-Resolution Image Generation.}
On COCO (Table \ref{tab:coco5k}), UPLiFT achieves a lower FID than CFM in a single step while reducing latency by 41\%.
UPLiFT is also close to matching the FID of SDXL, which natively generates 1024 images at much slower speeds.
On reLAION (Table \ref{tab:laion5k}), UPLiFT achieves strong performance and faster speeds than CFM, with our 40-step configuration producing the best CLIP, FID, and patch-FID, while maintaining significantly lower latency.
These results demonstrate UPLiFT as a strong choice for efficient generative feature upscaling.

\myparagraph{Image Super-Resolution.}
We evaluate UPLiFT for $4{\times}$ image super-resolution via latent-space-upsampling, comparing against two lightweight baselines: bilinear and nearest neighbors.
Note that the baseline CFM models for this task are trained specifically for the datasets tested, while UPLiFT uses the same general-purpose model from the previous section.
As shown in Table \ref{tab:super-res}, UPLiFT achieves competitive visual quality using just two iterative steps, with an inference time of only 271 milliseconds per image.
Remarkably, UPLiFT is just 13\% slower than bilinear upsampling in latent space, yet delivers an order-of-magnitude improvement in super-resolution quality.
These results highlight UPLiFT’s efficiency and generalization capabilities.

%% file: sec/5_conclusion.tex
\section{Conclusion}
\label{sec:conclusion}

In this work, we have presented UPLiFT, an efficient feature-upsampling method to create high-resolution feature maps from low-resolution features of pre-trained visual backbones. We show that UPLiFT produces high-performing pixel-dense features, and it does so with lower inference costs than existing comparable methods. UPLiFT achieves this through an iterative, convolutional upsampling architecture, without need for expensive cross-attention. Instead, we propose an efficient Local Attender operator, which succeeds in maintaining semantic consistency of features for minimal extra computational cost. Finally, we show that UPLiFT achieves state-of-the-art performance in a range of tasks, including both predictive and generative domains. Our code and UPLiFT models will be released at time of publication. We hope that UPLiFT will help to improve the efficiency and practicality of deep models for dense visual tasks.

\myparagraph{Acknowledgments.}
This work was partially supported by NSF CAREER Award (\#2238769) to AS. The U.S. Government is authorized to reproduce and distribute reprints for Governmental purposes notwithstanding any copyright annotation thereon. The authors acknowledge UMD’s supercomputing resources made available for conducting this research. The views and conclusions contained herein are those of the authors and should not be interpreted as necessarily representing the official policies or endorsements, either expressed or implied, of NSF or the U.S. Government.

%% file: sec_supp/A_predictive.tex
\section{Additional Details for Predictive Tasks}
\label{supp:predictive}

\subsection{Baseline Methods}

In this work, we focus on comparing UPLiFT with other task-agnostic feature upsamplers, which have grown in popularity in recent years. Such methods take existing pre-trained feature extractors, and provide an upsampler add-on to provide dense, powerful features directly out of the box. We compare with the following methods:

\myparagraph{FeatUp \cite{fu2024featup}:}
We compare with the JBU FeatUp variant, which is an iterative upsampler that performs $16\times$ upsampling with a stack of four modified Joint Bilinear Upsamplers \cite{kopf2007joint}. While the implicit variant of FeatUp produces impressive results, it has been shown by works like \cite{huang2025loftup} that this approach is too slow to be practical for large-scale evaluations, with functional inference speeds over $500\times$ slower than comparable methods. We use the official FeatUp JBU model distributed by \cite{fu2024featup} trained for DINOv2-S/14 without backbone normalization.

\myparagraph{LiFT \cite{suri2024lift}:}
LiFT is also an iterative upsampling method, which in its base form performs $2\times$ upsampling. The same LiFT model can also be applied iteratively four times to perform $16\times$ upsampling. However, \cite{couairon2025jafar} has shown that this iterative upsampling can lead to semantic drift and degraded features. For this reason, we follow the example of \cite{couairon2025jafar} and present LiFT in two configurations: ``LiFT'' runs the model four times for $16\times$ upsampling and ``LiFT-$2\times$'' runs the model only once, then follows this with bilinear upsampling to pixel scale. As there is not an official LiFT model for DINOv2-S/14, we train one for this work.

\myparagraph{LoftUp \cite{huang2025loftup}:}
LoftUp uses cross-attention to directly upsample low-resolution backbone features to pixel-scale features, with a $14\times$ upsampling step. We compare with the official LoftUp for DINOv2-S/14 model distributed by \cite{huang2025loftup}. We note that this LoftUp model was originally trained with the DINOv2-S/14 backbone distributed by \cite{oquab2023dinov2}, while other recent methods (ours included) use the version distributed by Hugging Face\footnote[1]{\urlA}. While these DINOv2 models are fundamentally the same model trained on the LVD-142M dataset and should produce similar results, we choose to conduct an additional test to determine if this has any impact on the performance of LoftUp. We perform evaluation twice with LoftUp, once with each version of DINOv2-S/14, and report the results in Table \ref{tab:bbvar}. We find that LoftUp's performance is nearly identical for both backbones, with only minor variations seen in the mIoU scores for VOC and Cityscapes. For this reason, we conclude that these minor variations in backbone distributions are not a significant confounding factor in our results, and we thus choose to present results for LoftUp with its original backbone for sake of fair representation.

\myparagraph{JAFAR \cite{couairon2025jafar}:}
JAFAR also uses a cross-attention-based approach to perform direct $14\times$ feature upsampling. We build on the evaluation protocols of JAFAR for segmentation and depth estimation, and we compare with the official JAFAR for DINOv2-S/14 model from \cite{couairon2025jafar}.

\input{tables_supp/loftup_comparison}

\myparagraph{AnyUp \cite{wimmer2025anyup}:} 
AnyUp proposes a modified version of the JAFAR architecture which is designed with an additional ``feature-agnostic layer'' that enables the model to generalize to different backbones at inference time. Note that AnyUp is still initially trained with features from a particular backbone for upsampling training. Moreover, the results of \cite{wimmer2025anyup} show that performing inference with a different backbone than was used in training leads to poorer downstream performance than using the same backbone for both training and evaluation, suggesting that training model-specific upsamplers is still preferable for performance. For this study, we compare with both AnyUp-v1 and AnyUp-v2 as distributed by \cite{wimmer2025anyup}. AnyUp-v2 differs from AnyUp-v1 in two key ways. First, it uses NATTEN \cite{hassani2023neighborhood} to improve the efficiency of its windowed attention operations. Second, it uses multiple proxy backbones during training, instead of a single backbone like AnyUp-v1. This use of multiple proxies is intended to improve generalization to new backbones, but it may also contribute to the slightly lower performance of AnyUp-v2 with DINOv2 as observed in Table \ref{tab:lp}.

\subsection{Upsampling Rates}

For our main evaluation, we use DINOv2-S/14 as the standard backbone, and run all methods in a configuration to produce pixel-dense features, which requires $14\times$ upsampling. For methods that perform $16\times$ upsampling, which includes our UPLiFT, we follow the example of \cite{couairon2025jafar} and simply over-upsample the features, and then downsample to pixel-resolution. We note that this puts UPLiFT at a computational efficiency disadvantage compared to cross-attention-based methods, but despite this, we still achieve faster inference than the cross-attention baselines. Moreover, we note that more recent self-supervised backbones, like DINOv3 \cite{simeoni2025dinov3}, have moved back to using patch size $16$, which makes $16\times$ upsampling the likely desired option for future research.

\subsection{Training Cost Comparison}

We briefly compare the training cost of UPLiFT to the recent state-of-the-art methods JAFAR \cite{couairon2025jafar}, AnyUp \cite{wimmer2025anyup}, and LoftUp \cite{huang2025loftup}. As discussed in Section \ref{sec:archtrain}, we train UPLiFT for one epoch on the ImageNet-1K dataset \cite{deng2009imagenet} using a multi-scale and multi-step training configuration. Under our final protocol, UPLiFT training with DINOv2-S/14 takes ${\sim}9$ hours on one A5000 GPU.
For comparison, when using the official training code of \cite{couairon2025jafar}, we find that JAFAR training takes ${\sim}3$ hours on one A5000. \cite{wimmer2025anyup} reports that AnyUp training takes ${\sim}5$ hours on one H100 GPU. The training protocols of AnyUp are based on those of JAFAR, though they use additional data augmentation with four random crops per image. While UPLiFT training does take a few hours longer than JAFAR and AnyUp, we believe all three are reasonably comparable as they can all be trained overnight with one moderate-strength GPU.
We also note that LoftUp training comes with a comparatively higher computational price. The official training code of \cite{huang2025loftup} recommends training with 4 GPUs, and when using 4 A5000s, we estimate that LoftUp would take ${\sim}50$ hours to train on a 1M image subset of SA1B \cite{kirillov2023segment} as recommended in \cite{huang2025loftup}.
Note that this estimate is only for the first training phase, and the additional self-distillation phase will add further time. Overall, the training cost for UPLiFT is comparable to JAFAR and AnyUp, and far lower than LoftUp.

\subsection{Local Attender Details}

To provide additional details on our implementation of Local Attender, we present pseudocode in Algorithm \ref{lapsuedo}. For sake of simplicity, we exclude the batch dimension from all notation. The primary operations used in Local Attender are matrix offsetting and element-wise matrix multiplication with broadcasting over placeholder dimensions. In broad terms, the steps are as follows. First, a $1{\times}1$ convolution is applied to Guide feature map $G$ to set its channel depth to $n$, the size of the Neighborhood. Softmax is applied to create the Attender Map, $A$. $G$ (and thus $A$) is larger than $V$ by an integer factor $c$ in the spatial dimensions, so we reshape $A$ to divide it into $c{\times}c$ cells. Then, the Value feature map $V$ is also processed.
Replication Padding is applied, and we then generate a set of offset feature maps, $V_\text{off}$, based on the Neighborhood positions. Element-wise matrix multiplication with broadcasting is then performed between $A$ and $V_\text{off}$, and the attention-weighted features are then summed. Finally, we concatenate the cells to produce the final upsampled feature map $\widetilde{V}$.

\subsection{UPLiFT Fast Mode}

As introduced in Section \ref{sec:upliftfast}, we present ``UPLIFT$_{\text{FAST}}$'' mode, which utilizes several additional inference-time optimizations to reduce UPLiFT's latency and memory usage. These optimizations are training-agnostic and can be applied to any pretrained UPLiFT model. The optimizations include the following:
\textbf{(1) BatchNorm folding:} we merge BatchNorm layers into the preceding convolutional layers, eliminating an entire layer per convolutional block at zero cost.
\textbf{(2) Channels last memory format:} we rearrange the tensor memory layout from NCHW to NHWC to align with the GPU's Tensor Cores' native format, avoiding format conversions during convolutions.
\textbf{(3) Fused Triton Local Attender kernel:} we implement the Local Attender as a single Triton kernel, replacing the pad/stack/softmax/sum sequence with a single launch and avoiding materialization of neighborhood tensor.
\textbf{(4) Graph compilation:} we apply \texttt{torch.compile}, which captures the model graph and fuses the remaining conv/norm/activation chains.
Combined, these optimizations significantly accelerate UPLiFT inference without harming downstream task performance. We further demonstrate this with additional speed and memory usage results in Appendix \ref{supp:results}.

\input{tables_supp/la_psuedo}

%% file: tables_supp/loftup_comparison.tex
\begin{table*}[t]
  \caption{\textbf{Evaluating the impact of minor backbone variations.} We present a comparison of baseline LoftUp \cite{huang2025loftup} running either with its original DINOv2-S/14 backbone from \cite{oquab2023dinov2}, or with an alternate distribution of DINOv2-S/14 provided by Hugging Face, which is commonly used in other upsampling works. Note that both models are theoretically the same model and only represent different distribution platforms. We find that there is minimal variation in the resulting performance, with the largest changes being in VOC mIoU and Cityscapes mIoU. Our UPLiFT surpasses LoftUp's performance for either backbone. We report results for LoftUp with its original backbone in the main work to provide a fair representation of the method.}
  \vspace{-0.12in}
  \label{tab:bbvar}
  \centering
  \resizebox{1.0\linewidth}{!}{
    \begin{tabular}{@{}lcc|cc|cc|cc|cc|cc@{}}
      \toprule
      \multicolumn{3}{@{}c}{} &
      \multicolumn{8}{c@{}}{Semantic Segmentation} &
      \multicolumn{2}{c@{}}{Depth Estimation} \\
      \cmidrule(r{2pt}){4-11}\cmidrule(l{2pt}){12-13}
      \multicolumn{3}{c}{Upsampler} &
      \multicolumn{2}{c}{COCO} &
      \multicolumn{2}{c}{VOC} &
      \multicolumn{2}{c}{ADE20K} &
      \multicolumn{2}{c}{Cityscapes} &
      \multicolumn{2}{c}{COCO} \\
      \cmidrule(r{2pt}){1-3}\cmidrule(l{2pt}r{2pt}){4-5}\cmidrule(l{2pt}r{2pt}){6-7}\cmidrule(l{2pt}r{2pt}){8-9}\cmidrule(l{2pt}r{2pt}){10-11}\cmidrule(l{2pt}){12-13}
      Method & Params\,(M) & Time\,(ms) &
      mIoU $\uparrow$ & Acc $\uparrow$ & mIoU $\uparrow$ & Acc $\uparrow$ &
      mIoU $\uparrow$ & Acc $\uparrow$ & mIoU $\uparrow$ & Acc $\uparrow$ &
      $\delta_1 \uparrow$ & RMSE $\downarrow$ \\
      \midrule
      LoftUp + Orig Backbone            & 4.3 & 223.5 & 62.19 & 81.35 & 84.63 & 96.33 & 42.16 & 75.79 & 62.09 & 93.55 & 58.69 & 0.68 \\
      LoftUp + HF Backbone            & 4.3 & 223.5 & 62.20 & 81.35 & 84.52 & 96.30 & 42.16 & 75.77 & 62.17 & 93.56 & 58.70 & 0.68 \\
      \midrule
      \textbf{UPLiFT}            & 0.8 & 79.4 & \textbf{62.55} & \textbf{81.57} & \textbf{85.21} & \textbf{96.51} & \textbf{42.97} & \textbf{76.00} & \textbf{65.38} & \textbf{94.41} & \textbf{61.16} & \textbf{0.63} \\
      \bottomrule
      \vspace{-0.25in}
    \end{tabular}
  }
\end{table*}

%% file: tables_supp/la_psuedo.tex
\begin{algorithm*}
  \centering
  \caption{$\widetilde{V} \gets \text{LocalAttender}(V,G|W,N)$}
  \label{lapsuedo}
  \begin{minipage}{2.05\columnwidth}
  \textbf{LocalAttender} upsamples ``Value" feature map $V$ using a high-res ``Guide" feature $G$ to apply local attention over Neighborhood $N$, which is defined as a collection of fixed integer offsets in 2D space.\\
  \hspace*{\algorithmicindent} \textbf{Inputs:} $V \in \mathbb{R}^{C_V \times H \times W}$ and $G \in \mathbb{R}^{C_G \times cH \times cW}$ where $c \in \mathbb{Z}$\\
  \hspace*{\algorithmicindent} \textbf{Outputs:} $\widetilde{V} \in \mathbb{R}^{C_V \times cH \times cW}$ \\
  \hspace*{\algorithmicindent} \textbf{Parameters:} $W \in \mathbb{R}^{C_G \times n}$ where $n = ||N||$ \\
  \hspace*{\algorithmicindent} \textbf{Hyperparameters:} $N = {o_1, o_2, ..., o_n}$ where $o_i \in \mathbb{Z}^2$
  \begin{algorithmic}[1]
    \State $A \gets \text{Conv}_{1\times1}(G,W)$
    \Comment{Attender Map $A$ now has shape $[n, cH, cW]$}
    \State $A \gets \text{Softmax}(A, \text{dim}=1)$
    \State $A \gets \text{DivideCells}(A, \text{size}=(c,c))$
    \Comment{$A$ now has shape $[n, H, c, W, c]$}
    \State $A \gets \text{ExpandDims}(A, \text{dims}=1)$
    \Comment{$A$ now has shape $[1, n, H, c, W, c]$}
    \State $V_{\text{pad}} \gets \text{ReplicationPad2d}(V)$
    \State $V_{\text{off}} \gets \text{Stack}(\text{ApplyOffset}(V_{\text{pad}},o_i)
    \text{~} \forall o_i \in N)$
    \Comment{$V_\text{off}$ now has shape $[C_V, n, H, W]$}
    \State $V_{\text{off}} \gets \text{ExpandDims}(V_{\text{off}}, \text{dims}=(4,6))$
    \Comment{$V_\text{off}$ now has shape $[C_V, n, H, 1, W, 1]$}
    \State $\widetilde{V} \gets \text{BroadcastMultiply}(A, V_{\text{off}})$
    \Comment{$\widetilde{V}$ now has shape $[C_V, n, H, c, W, c]$}
    \State $\widetilde{V} \gets \text{Sum}(\widetilde{V}, \text{dim}=2)$
    \Comment{$\widetilde{V}$ now has shape $[C_V, H, c, W, c]$}
    \State $\widetilde{V} \gets \text{ConcatCells}(\widetilde{V})$
    \Comment{$\widetilde{V}$ now has shape $[C_V, cH, cW]$}
  \end{algorithmic}
  \end{minipage}
  \vspace{0.1in}
\end{algorithm*}

%% file: sec_supp/B_generative.tex
\section{Additional Details for Generative Tasks}
\label{supp:generative}

\subsection{UPLiFT for VAE Features}

We describe additional details of our UPLiFT model designed for VAE feature upsampling and generative tasks.

\myparagraph{UPLiFT size for VAE.}
We empirically find that small parameter count upsampling models, like those demonstrated to be effective for predictive downstream tasks, are insufficient for generative downstream tasks. We illustrate this in Figure \ref{fig:genabl}, where we present a comparison with a small UPLiFT model, which has roughly $2.8M$ parameters and is trained using the same training protocols as our main model. In this comparison, we see that the resulting model produces blurry, low quality upsampling, and is not able to capture high-frequency information.
We theorize that the necessary model capacity to train an effective feature upsampler for generative tasks is intrinsically larger than the necessary size for predictive tasks. We base this theory on the intuition that predictive tasks are about narrowing down information into a compressed understanding, which can be represented more compactly with a smaller network, while generative tasks require a high level of detail over a diverse range of visual textures and patterns to achieve high quality. For this reason, we train a larger UPLiFT model for generative tasks, increasing the number of layers in both the encoder and decoder module and increasing the channels per layer. This larger size UPLiFT has $53.5M$ parameters, which still makes it significantly smaller than the compared CFM \cite{schusterbauer2024fmboost} models, which have $113M$ or $306M$ parameters. 

\input{figures_supp/gen_ablation_fig}

\myparagraph{Refiner Block.}
We find that it is beneficial to introduce an additional ``Refiner Block'' after the Local Attender module, which is designed to realign the output features with the distribution expected by the VAE decoder. We demonstrate the importance of the refiner block by ablating it in Figure \ref{fig:genabl} (Right). Without the refiner block, the upsampled images have significant blocky artifacts, which are likely the result of the strict, linear-combination feature upsampling approach. We note that our Local Attender module, and the cross-attention modules used by recent works, act as a form of strong feature regularization, which ensures that the features output by the upsampler maintain a similar distribution to the original input distribution, which, in theory should also match the distribution expected by the VAE decoder. However, through testing, we find that this design may be too restrictive for a generative context. For this reason, we introduce a post-attender ``Refiner Block'' which gives UPLiFT an opportunity to improve the final features.

\myparagraph{Noise Layers and Augmentation.}
We follow the example of \cite{schusterbauer2024fmboost} and concatenate additional gaussian random noise channel inputs at several points in the UPLiFT Encoder module when upsampling VAE features. These additional noise channels are provided to help seed the generation of high resolution details. We also apply gaussian noise augmentation to the input latent representation, following \cite{schusterbauer2024fmboost}.

\myparagraph{Color Correction.}
When training UPLiFT for VAE, all our loss terms are computed in the latent feature space, which improves the efficiency of training by removing the need to decode high-resolution images.
Our UPLiFT training is effective at minimizing the $L2$ distance of upsampled features in latent space, however, we find that small perturbations in latent space can lead to slight color shifts after decoding. While this could likely be addressed through the use of pixel-space loss terms, this would greatly increase training costs. Instead, we introduce a simple color correction module after the VAE Decoder at inference time. This module computes the per-color-channel means for the low-resolution input image and the high-resolution output image, and subtracts the difference vector from the output image to realign the color mean. We find that this simple module is sufficient to remove any minor color shifts.

\myparagraph{Layer Normalization.}
Finally, for our larger VAE UPLiFT model, we find it is beneficial to replace the Batch Norm~\cite{ioffe2015batch} layers with Layer Norm~\cite{ba2016layer} instead, as it provides better numerical stability for the deeper architecture. When using Batch Norm, we sometimes observe color blob artifacts in the upsampled images, which are a consequence of numerical instability in the model. After replacing the Batch Norm operations with Layer Norm, these artifacts no longer occur.

\subsection{Experimental Methods for Generative Tasks}

We summarize additional key details of our experimental protocols for generative tasks. 

\myparagraph{Diffusion Upsampling.}
We evaluate UPLiFT's ability to upscale diffusion features on COCO 2014 and reLAION-400M following the protocols of \cite{schusterbauer2024fmboost}.
On COCO, we randomly sample 5k caption-image pairs.
On reLAION, we randomly sample 5k images with a minimum resolution of $1024{\times}1024$.
While \cite{schusterbauer2024fmboost} used the now deprecated LAION dataset, our sample is sufficiently similar, and our computed FID with LCM-LoRA SDXL matches that of CFM.
During inference, we generate latents with Stable Diffusion 1.5\,\cite{rombach2021highresolution} for each sampled caption, and then decode them to generate 5k $512{\times}512$ images.
We apply UPLiFT to produce $1024{\times}1024$ images. Note that UPLiFT leverages the image created from the decoded low-resolution latents to guide feature upsampling, similar to CFM, which must decode the image as part of the Pixel-Space Upsampling (PSU) method. However, UPLiFT does not require re-encoding a bilinearly upsampled image like CFM does, which gives an additional efficiency gain to UPLiFT. 

The default precision of \texttt{float32} is used for all diffusion processes, and each model utilizes its default scheduler with the specified number of steps. 
We also compare with SDXL\,\cite{podell2023sdxl} and LCM-LoRA SDXL\,\cite{luo2023lcm}, which natively generate at $1024{\times}1024$, to provide an additional point of reference for visual quality from high-resolution latents. SDXL can be viewed as an `upper-bound' oracle for native megapixel generation, while LCM-LoRA SDXL provides a point of reference of a low-latency distilled model at megapixel native generation. Please note that SD 1.5 cannot natively generate latents for images at resolutions different from $512{\times}512$ without significant quality degradation or alterations. 
\cite{schusterbauer2024fmboost} includes experiments upscaling $256{\times}256$ to $1024{\times}1024$ by specially fine-tuning a Stable Diffusion 1.5 model at a lower resolution.
For our tests, we choose to prioritize experiments using only the official Stable Diffusion 1.5 model generating latents for $512{\times}512$ images.
Utilizing the official code of \cite{schusterbauer2024fmboost}, we compute FID and patch-FID, which splits the image into four random patches sized $512{\times}512$.
For CLIP score, we use the Python package \texttt{clip\_score} with \texttt{clip-vit-base-patch32}.

\input{figures_supp/neighbors}

\myparagraph{Super-Resolution.}
Following CFM \citep{schusterbauer2024fmboost}, we perform super-resolution evaluations on two datasets: FacesHQ and LHQ.
We randomly sample 5k images from the LHQ dataset and the joint combination of CelebA-HQ and FFHQ datasets, called FacesHQ.
For each image, we perform a $1024{\times}1024$ center crop, followed by a bilinear downsample to $256{\times}256$.
Finally, we apply UPLiFT to recover the full sized image.
To do so, we first use Stable Diffusion 1.5’s VAE to encode the image into latent space.
Then we apply the UPLiFT module with 2 iterations, resulting in a $4{\times}$ latent, which is decoded using the same VAE's decoder back into pixel space.
Performance is measured using the evaluation scripts of \cite{schusterbauer2024fmboost}, which track SSIM\,\cite{wang2004image}, PSNR, FID, and patch-FID.

\myparagraph{Latency Timing.}
To ensure fair comparison with \cite{schusterbauer2024fmboost},  we measure latency on a single NVIDIA A100-SXM4-80GB across both experiments. 
We use a batch size of 1 for experiments on reLAION, and 4 for experiments on COCO to be consistent with \cite{schusterbauer2024fmboost}. All experiments utilize the \texttt{torch.compile} module, and measurements are conducted with 3 warm-up batches. The final latency is averaged across 10 subsequent batches.

%% file: figures_supp/gen_ablation_fig.tex
\begin{figure*}[t]
    \centering
    \includegraphics[width=1.0\linewidth]{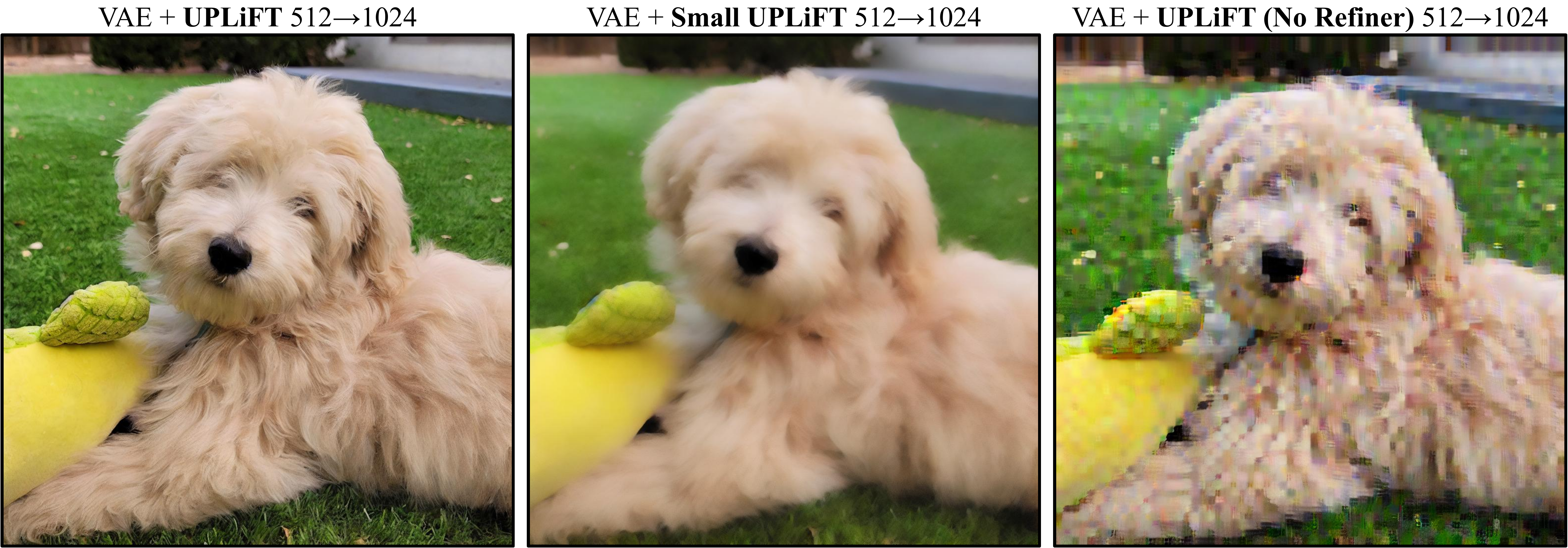}
    \vspace{-0.27in}
    \caption{\textbf{Visual ablation of design choices for VAE UPLiFT.} (Left) UPLiFT achieves high quality 512$\to$1024 upsampling with a larger parameter count model. (Middle) A smaller-scale UPLiFT has insufficient capacity to upsample all high-frequency information and produces blurry results. (Right) Ablation of the Refiner Block leads to blocky artifacts in upsampled images.}
    \label{fig:genabl}
\end{figure*}

%% file: figures_supp/neighbors.tex
\begin{figure*}[t]
    \centering
    \includegraphics[width=0.8\linewidth]{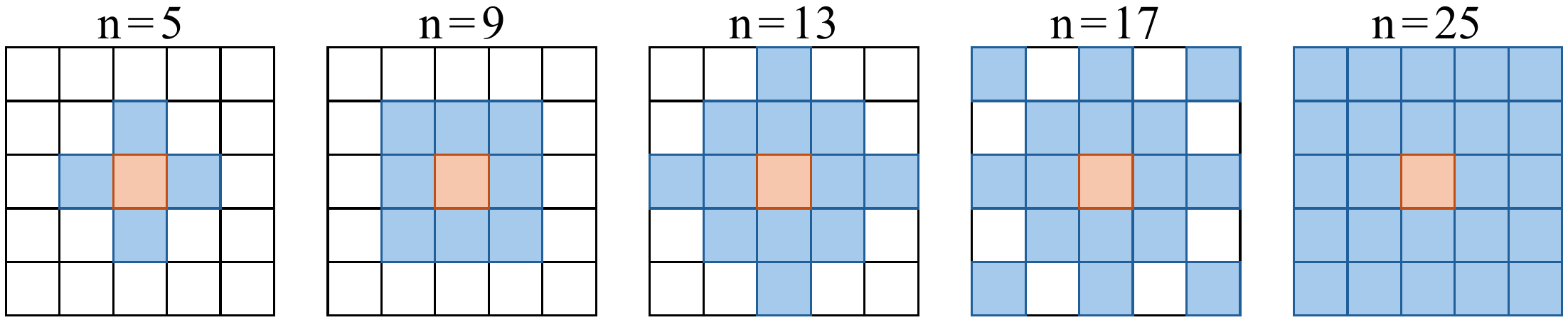}
    \vspace{-0.1in}
    \captionsetup{width=0.8\linewidth}
    \caption{\textbf{Local Attender neighborhood designs.} We visualize the neighborhood designs tested with our Local Attender module. The center token (orange) is always included in the neighborhood, and the blue tokens represent the offset relative local positions that are included in feature pooling through local attention.}
    \label{fig:neighbors}
\end{figure*}

%% file: sec_supp/C_uplift_ablations.tex
\section{Ablations of UPLiFT Design Choices}
\label{supp:ablations}

\subsection{Training Depth}
\label{supp:ablationsdepth}

As discussed in Section \ref{sec:archtrain}, we train UPLiFT with a multi-depth, multi-step feature reconstruction training objective, where the level of downsampling and the number of upsampling steps depends on the training depth, $d$. Here we present an ablation over $d$, where we train our UPLiFT model with different training depth configurations, including individual depths, or multiple concurrent depths. We use semantic segmentation on COCO and VOC with pixel-dense features as our evaluation tasks for this ablation. The results are summarized in Table \ref{tab:ablatedepth}.

\input{tables_supp/ablate_depth}

When training with only one depth level (rows $1{-}4)$, better performance is achieved with shallower depths. VOC has the best performance with $d{=}1$, and COCO has slightly better performance with $d{=}2$. We believe this occurs because shallower depths give the UPLiFT Encoder a large input image, which allows it to learn more about high-frequency image information. For $d{=}3$, we see a slight drop in performance and for $d{=}4$ we see a major drop in performance. Under $d{=}4$, the $448{\times}448$ image is downsampled to $24{\times}24$ which yields only a $2{\times}2$ token grid. It seems this level of downsampling is too severe for effective learning.

Next, we find that training with multiple depths concurrently can enhance performance. As shown by row $5$, training with $d\in\{1,2\}$ concurrently leads to superior performance on both COCO and VOC. This effect is further enhanced by adding in $d{=}3$. However, the addition of $d{=}4$ becomes detrimental for multi-depth training, which again indicates that downsampling the input too severely hinders learning. From this, we find that training with a multi-depth configuration with $d\in\{1,2,3\}$ yields the best results, so we use this training configuration in our predictive task experiments. Note that for our VAE UPLiFT model, we train with a maximum resolution of $1024{\times}1024$ in following \cite{schusterbauer2024fmboost}. For this larger image size, we find that $d\in\{1,2,3,4\}$ works to good effect.

\input{tables_supp/ablate_neighbor}

\subsection{Local Attender Neighborhood}

We propose a new Local Attender module, which uses a relatively-defined, variably-sized local neighborhood to perform local attention pooling in linear time. In this section, we present additional experiments with different neighborhood sizes and shapes. We limit the neighborhood size to a maximum of a $5{\times}5$. We present an illustration of the neighborhood sizes and shapes tested in Figure \ref{fig:neighbors}. In addition, we present results for an UPLiFT model with the Local Attender module ablated. All models are trained for one epoch on ImageNet-1k, with $448{\times}448$ max image size, like the main work. Results are summarized in Table \ref{tab:ablatelocatt}. 

First, from row $1$, we see that the removal of the Local Attender module significantly harms performance, which indicates that our Local Attender is essential for maintaining effective feature upsampling through multiple steps. Next, in row $2$, we test $n{=}5$ with what we consider to be the smallest viable neighborhood, which consists of only the current token and its immediately touching neighbors. Already for $n{=}5$ we see strong performance in both datasets, though this performance continues to improve as we expand the neighborhood to $9$, $13$, and $17$ neighbors. The $n{=}25$ configuration, which compared to $n{=}17$ adds on several additional neighbors along the second layer of offsets, only leads to roughly equal performance in VOC and slightly lower performance in COCO. This result indicates that the addition of further neighbors may not be beneficial if they are added in intermediate places between existing neighbors. Overall, our best performance is achieved with the star-shaped $n{=}17$ neighborhood design, so we use this as our Local Attender neighborhood for all our experiments in the main work.

\input{tables_supp/ablate_enc_dec}
\input{tables_supp/ablate_wca}
\input{tables_supp/other_backbones}

\input{tables_supp/dinov3_results}

\subsection{UPLiFT Encoder and Decoder Designs}

In this section, we provide additional details of our UPLiFT Encoder and Decoder modules, and we present ablations testing larger and smaller variations of each module. Our UPLiFT Decoder module does the primary work of upsampling the low resolution backbone features, while the UPLiFT Encoder extracts shallow, high-res features to help guide feature upsampling. Unlike LiFT, which runs its encoder module iteratively after each upsampling step, we use a single-encoder-pass approach, where the encoder module is only run once with the original resolution input image. Our encoder maintains the same spatial resolution as the input image, and nearest neighbor downsampling is applied to the resulting features before guiding each upsampling step. Our Encoder and Decoder modules are both lightweight, convolutional networks, with 10 and 6 conv layers respectively, along with batchnorm and ReLU operations. The Decoder also includes a deconv layer to increase the feature resolution ${\times}2$. Our official UPLiFT repository (\url{https://github.com/mwalmer-umd/UPLiFT/}) includes code and configuration files with full details for both of these modules.

We further demonstrate the optimality of our final UPLiFT design by comparing it with variations with larger and smaller encoders and decoders. These variations are creating by halving or doubling the channel depth of each convolutional layer. We additionally present an ablation where the encoder module is removed completely.
As shown in Table \ref{tab:ablateencdec}, we find that the smaller modules underperform, while the larger ones increase cost (doubled parameters/slower speed) with minimal performance gain. In addition, fully ablating the encoder module significantly harms COCO and VOC performance. Thus, our proposed UPLiFT design offers the best performance-efficiency trade-off compared to the other configurations.

\subsection{Comparison with Windowed Self-Attention}

We present an ablation testing the effectiveness of our Local Attender module in comparison with Windowed Cross-Attention (WCA), which may also be referred to as Neighborhood Attention \cite{hassani2023neighborhood}. Like Local Attender, WCA can provide a mechanism to pool local features to produce an upsampled feature map that adheres to the distribution of the original backbone map. The concurrent work AnyUp leverages WCA in this way.
When implemented efficiently, WCA can also avoid the quadratic time scaling of full cross-attention. However, we believe our Local Attender has an additional advantage: Local Attender re-formulates the local attentional operations using consistent, relatively-defined local-offsets, which we theorize provides stronger inductive biases to guide learning.
To demonstrate this, we train several UPLiFT models which use WCA in place of Local Attender. We use the efficient implementation of NATTEN \cite{hassani2023neighborhood} with a window size of $5{\times}5$ to provide a level comparison.
We test WCA with several attention head counts, ranging from 1 to 12. From our result in Table \ref{tab:ablatewca}, we find that WCA with one head has significantly lower performance. Running WCA with 4 or 8 heads achieves performance more comparable to the baseline method JAFAR, while 12 heads is actually detrimental. However, our UPLiFT with Local Attender has better performance in both metrics and datasets over all variations using WCA. Overall, these results demonstrate the potential strengths of our Local Attender design over Windowed Cross-Attention for use in feature upsampling.

%% file: tables_supp/ablate_depth.tex
\begin{table}[t]
  \renewcommand{\tabcolsep}{9pt}
  \caption{\textbf{Ablation of training depth configurations.} We test COCO and VOC segmentation for different training depth levels, including individual and concurrent depth configurations.}
  \vspace{-0.12in}
  \label{tab:ablatedepth}
  \centering
  \resizebox{1.0\linewidth}{!}{
    \begin{tabular}{@{}l|cc|cc@{}}
      \toprule
      \multicolumn{1}{@{}c}{} &
      \multicolumn{2}{c@{}}{COCO} &
      \multicolumn{2}{c@{}}{VOC} \\
      \cmidrule(r{2pt}){2-3}\cmidrule(r{2pt}){4-5}
      Depth & mIoU $\uparrow$ & Acc $\uparrow$ & mIoU $\uparrow$ & Acc $\uparrow$ \\
      \midrule
      $d = 1$ & 62.26 & 81.34 & 85.03 & 96.42 \\
      $d = 2$ & 62.34 & 81.43 & 84.89 & 96.40 \\
      $d = 3$ & 62.24 & 81.34 & 84.67 & 96.33 \\
      $d = 4$ & 60.82 & 80.30 & 81.42 & 95.42 \\
      \midrule
      $d \in \{1, 2\}$ & \underline{62.46} & \underline{81.50} & \underline{85.04} & \underline{96.44} \\
      $d \in \{1, 2, 3\}$ & \textbf{62.55} & \textbf{81.57} & \textbf{85.21} & \textbf{96.51} \\
      $d \in \{1, 2, 3, 4\}$ & 62.36 & 81.45 & 84.97 & 96.42 \\
      \bottomrule
    \end{tabular}
  }
\end{table}

%% file: tables_supp/ablate_neighbor.tex
\begin{table}[t]
  \renewcommand{\tabcolsep}{11.2pt}
  \renewcommand{\arraystretch}{1.125}
  \caption{\textbf{Ablation of Local Attender neighborhood sizes.} We compare UPLiFT performance with Local Attender modules with different neighborhood patterns, or without the Local Attender.}
  \vspace{-0.12in}
  \label{tab:ablatelocatt}
  \centering
  \resizebox{1.0\linewidth}{!}{
    \begin{tabular}{@{}l|cc|cc@{}}
      \toprule
      \multicolumn{1}{@{}c}{} &
      \multicolumn{2}{c@{}}{COCO} &
      \multicolumn{2}{c@{}}{VOC} \\
      \cmidrule(r{2pt}){2-3}\cmidrule(r{2pt}){4-5}
      Neighbors & mIoU $\uparrow$ & Acc $\uparrow$ & mIoU $\uparrow$ & Acc $\uparrow$ \\
      \midrule
      No LA & 47.52 & 73.06 & 64.86 & 91.37 \\
      \midrule
      $n = 5$  & 62.15 & 81.30 & 84.80 & 84.80 \\
      $n = 9$  & 62.33 & 81.42 & 84.99 & 96.44 \\
      $n = 13$ & \underline{62.51} & \underline{81.53} & 85.10 & 96.47 \\
      $n = 17$ & \textbf{62.55} & \textbf{81.57}& \underline{85.21} & \textbf{96.51} \\
      $n = 25$ & 62.46 & 81.50 & \textbf{85.22} & \underline{96.48} \\
      \bottomrule
    \end{tabular}
  }
\end{table}

%% file: tables_supp/ablate_enc_dec.tex
\begin{table}[t]
  \renewcommand{\tabcolsep}{5pt}
  \renewcommand{\arraystretch}{1.0}
  \caption{\textbf{Ablation of Encoder and Decoder Size.} We compare our final UPLiFT design to variations with smaller or larger encoders and decoders or alternately no encoder module.}
  \vspace{-0.12in}
  \label{tab:ablateencdec}
  \centering
  \resizebox{1.0\linewidth}{!}{
    \begin{tabular}{@{}lcc|cc|cc@{}}
      \toprule
      \multicolumn{3}{@{}c}{Upsampler} &
      \multicolumn{2}{c@{}}{COCO~~} &
      \multicolumn{2}{c@{}}{VOC} \\
      \cmidrule(r{2pt}){1-3}\cmidrule(r{2pt}){4-5}\cmidrule(r{2pt}){6-7}
      Model & Params (M) & Time (ms) & mIoU $\uparrow$ & Acc $\uparrow$ & mIoU $\uparrow$ & Acc $\uparrow$ \\
      \midrule
      Enc$_{None}$  & 0.36 & 64.6 & 61.81 & 81.02 & 84.19 & 96.21 \\
      Enc$_{Small}$ & 0.50 & 71.4 & 62.48 & 81.51 & \underline{85.21} & 96.46 \\
      Enc$_{Large}$ & 1.79 & 98.7 & 62.51 & 81.54 & \textbf{85.22} & \underline{96.48} \\
      \midrule
      Dec$_{Small}$ & 0.50 & 75.9 & 62.37 & 81.43 & 84.91 & 96.41 \\
      Dec$_{Large}$ & 1.57 & 88.3 & \textbf{62.58} & \textbf{81.60} & \textbf{85.22} & \underline{96.48} \\
      \midrule
      UPLiFT        & 0.79 & 79.4 & \underline{62.55} & \underline{81.57} & \underline{85.21} & \textbf{96.51} \\
      \bottomrule
    \end{tabular}
  }
\end{table}

%% file: tables_supp/ablate_wca.tex
\begin{table}[t]
  \renewcommand{\tabcolsep}{4pt}
  \renewcommand{\arraystretch}{1.0}
  \caption{\textbf{Local Attender \vs Windowed Cross Attention.} We present an ablation using Windowed Cross Attention (WCA) in place of our Local Attender (LA) module. Local Attender results in better performance and efficiency over WCA.}
  \vspace{-0.12in}
  \label{tab:ablatewca}
  \centering
  \resizebox{1.0\linewidth}{!}{
    \begin{tabular}{@{}lccc|cc|cc@{}}
      \toprule
      \multicolumn{4}{@{}c}{Upsampler} &
      \multicolumn{2}{c@{}}{COCO~~~} &
      \multicolumn{2}{c@{}}{VOC} \\
      \cmidrule(r{2pt}){1-4}\cmidrule(r{2pt}){5-6}\cmidrule(r{2pt}){7-8}
      Att. & Heads & Params (M) & Time (ms) & mIoU $\uparrow$ & Acc $\uparrow$ & mIoU $\uparrow$ & Acc $\uparrow$ \\
      \midrule
      WCA   & 1 & 0.96 & 87.4 & 61.52 & 80.92 & 84.06 & 96.15 \\
      WCA   & 4 & 0.96 & 90.0 & \underline{61.59} & \underline{81.04} & 84.12 & \underline{96.21} \\
      WCA   & 8 & 0.96 & 91.6 & 61.53 & 80.99 & \underline{84.25} & 96.19 \\
      WCA   & 12 & 0.96 & 93.8 & 61.35 & 80.84 & 83.98 & 96.13 \\
      \midrule
      LA    & - & 0.79 & 79.4 & \textbf{62.55} & \textbf{81.57} & \textbf{85.21} & \textbf{96.51} \\
      \bottomrule
    \end{tabular}
  }
\end{table}

%% file: tables_supp/other_backbones.tex
\begin{table}[t]
  \renewcommand{\tabcolsep}{10.5pt}
  \renewcommand{\arraystretch}{1.0}
  \caption{\textbf{UPLiFT with Additional Non-ViT Backbones.} We demonstrate the effectiveness of UPLiFT with additional convolution-based backbone's features.}
  \vspace{-0.12in}
  \label{tab:otherbackbones}
  \centering
  \resizebox{1.0\linewidth}{!}{
    \begin{tabular}{@{}ll|cc|cc@{}}
      \toprule
      \multicolumn{2}{@{}c}{Features} &
      \multicolumn{2}{c@{}}{COCO~~~~} &
      \multicolumn{2}{c@{}}{VOC} \\
      \cmidrule(r{2pt}){1-2}\cmidrule(r{2pt}){3-4}\cmidrule(r{2pt}){5-6}
      Backbone & Upsampler & mIoU $\uparrow$ & Acc $\uparrow$ & mIoU $\uparrow$ & Acc $\uparrow$ \\
      \midrule
      ResNet-50 & Bilinear & 40.26 & 65.28 & 49.87 & 85.72 \\
      ResNet-50 & UPLiFT & \textbf{43.90} & \textbf{68.81} & \textbf{54.75} & \textbf{86.94} \\
      \midrule
      ConvNeXt-S & Bilinear & 45.76 & 68.15 & 57.71 & 87.14 \\
      ConvNeXt-S & UPLiFT & \textbf{47.14} & \textbf{69.37} & \textbf{59.48} & \textbf{87.48} \\
      \bottomrule
    \end{tabular}
  }
  \vspace{-0.1in}
\end{table}

%% file: tables_supp/dinov3_results.tex
\begin{table*}[t]
  \caption{\textbf{Semantic Segmentation with DINOv3 and UPLiFT.} We measure segmentation performance on COCO and VOC for DINOv2-S/14 \vs DINOv3-S+/16. We compare against JAFAR and AnyUp, which either report results with this DINOv3 backbone or have provided an official model for it \cite{couairon2025jafar}. We find UPLiFT gives the best performance for all metrics, datasets, and backbones. $^\dagger$Row adapted from \cite{wimmer2025anyup}.}
  \renewcommand{\tabcolsep}{12pt}
  \vspace{-0.12in}
  \label{tab:dinov3}
  \centering
  \resizebox{1.0\linewidth}{!}{
    \begin{tabular}{@{}lc|cc|cc|cc|cc@{}}
      \toprule
      \multicolumn{2}{@{}c}{} &
      \multicolumn{4}{c@{}}{DINOv2-S/14} &
      \multicolumn{4}{c@{}}{DINOv3-S+/16} \\
      \cmidrule(r{2pt}){3-6}\cmidrule(l{2pt}){7-10}
      \multicolumn{2}{c}{Upsampler} &
      \multicolumn{2}{c}{COCO} &
      \multicolumn{2}{c}{VOC} &
      \multicolumn{2}{c}{COCO} &
      \multicolumn{2}{c}{VOC}\\
      \cmidrule(r{2pt}){1-2}\cmidrule(l{2pt}r{2pt}){3-4}\cmidrule(l{2pt}r{2pt}){5-6}\cmidrule(l{2pt}r{2pt}){7-8}\cmidrule(l{2pt}r{2pt}){9-10}
      Method & Params\,(M) &
      mIoU $\uparrow$ & Acc $\uparrow$ & mIoU $\uparrow$ & Acc $\uparrow$ &
      mIoU $\uparrow$ & Acc $\uparrow$ & mIoU $\uparrow$ & Acc $\uparrow$ \\
      \midrule
      JAFAR & 0.7 & 61.71 & 81.01 & \underline{84.38} & \underline{96.22} & 62.47 & 81.50 & \underline{83.05} & \underline{95.99} \\
      AnyUp$^\dagger$ & 0.9 & \underline{62.16} & \underline{81.37} & 84.00 & 96.19 &\underline{62.99} & \underline{81.84} & -- & -- \\
      \midrule
      \textbf{UPLiFT} & 0.8 & \textbf{62.55} & \textbf{81.57} & \textbf{85.21} & \textbf{96.51} & \textbf{63.74} & \textbf{82.27} & \textbf{84.72} & \textbf{96.55} \\
      \bottomrule
    \end{tabular}
  }
  \vspace{-0.05in}
\end{table*}

%% file: sec_supp/D_results.tex
\section{Additional Results and Visualizations}
\label{supp:results}

\subsection{UPLiFT for Convolutional Backbones}

To demonstrate the generality of UPLiFT, Table \ref{tab:otherbackbones} presents results for two additional non-ViT timm backbones with different training protocols: ResNet-50 with YFCC100M ssl-pretraining \& IN1k finetuning, and ConvNeXt-S with IN22k pretraining \& IN1k finetuning. We use bilinear feature upsampling as a baseline. UPLiFT yields better performance for both datasets and metrics. We would also note that the SD1.5 VAE backbone used in Section \ref{sec:expgen} is primarily convolution-based. Overall, these results demonstrate that our UPLiFT approach is effective for both ViT-based and convolution-based backbones and features.

\subsection{UPLiFT for DINOv3}

In our primary results in Section \ref{sec:exppred}, we follow the example of recent feature upsampling works and use DINOv2-S/14\,\cite{oquab2023dinov2} as our primary backbone for assessing UPLiFT and comparing it with baseline works. However, as of writing, DINOv3\,\cite{simeoni2025dinov3} has recently been published, and we expect future work in feature upsampling will likely shift to focus on this family of backbones. For this reason, we present additional results for Semantic Segmentation using DINOv3-S+/16. We compare with JAFAR\,\cite{couairon2025jafar}, which has released an official model for this backbone on their repository, and with AnyUp\,\cite{wimmer2025anyup}, which has published results for DINOv3 on COCO in their work. While we would additionally like to compare with LoftUp\,\cite{huang2025loftup}, no official DINOv3 LoftUp upsampler has been published as of writing, nor has official training code been made publicly available. We present results for COCO and VOC in Table \ref{tab:dinov3}, alongside results for DINOv2-S/14 for comparison. We see that the DINOv3 features lead to stronger performance in COCO, with improved results for DINOv3 with UPLiFT. However, DINOv3 does not necessarily yield stronger performance in VOC, as is seen for both UPLiFT and JAFAR. However, for both the DINOv2 and DINOv3 backbones, UPLiFT achieves the best semantic segmentation performance for both metrics and datasets.

\subsection{Qualitative Comparisons with Baselines}

\input{figures_supp/pca_comp}

Following the example of prior works \cite{fu2024featup, couairon2025jafar, wimmer2025anyup, huang2025loftup}, we present a qualitative comparison of our upsampled features with baseline methods using Principal Component Analysis (PCA).
As shown in Figure \ref{fig:pcacomp}, UPLiFT and the other recent methods are effective at maintaining semantic consistency for key objects, as indicated by consistent coloration with the base features. LiFT's upsampled features show signs of visual degradation, which we examine more in Appendix \ref{sec:suppsemdrift}.
Qualitatively, UPLiFT's features are comparable to the recent cross-attention-based methods JAFAR, AnyUp, and LoftUp. We note that UPLiFT features can sometimes appear smoother, which is likely a result of our iterative feature growing method.
In some cases, JAFAR, AnyUp, and LoftUp appear sharper, but this sharpness often comes with noisier edges. This can also lead to feature-bleed, where features from unrelated background regions are found in foreground objects.
This effect can be seen in row 2 for JAFAR and AnyUp and in row 3 for LoftUp.
UPLiFT's focus on iterative feature growing with local neighborhoods tends to produce more locally consistent features, which is likely beneficial for downstream tasks like segmentation.

We also note that DINOv2's underlying features tend to have noisy artifacts in smooth background regions, like the sky, as shown in row 2.
While most methods preserve these artifacts in some form after upsampling, LoftUp is the only method that actively removes them. This is a result of LoftUp's training protocol, which uses additional pseudo-supervision from SAM \cite{kirillov2023segment} to refine and remove these artifacts from the training data. While the removal of these artifacts may look better qualitatively, it is difficult to judge if this is practically beneficial, as it does cause LoftUp to deviate more significantly from the original backbone's feature distribution. In addition, we note that the recently released DINOv3 \cite{simeoni2025dinov3} exhibits fewer background artifacts compared to DINOv2, so actively removing artifacts from training data may be a lower priority for future models. In summary, these PCA visualizations help to reveal differing properties of the features created by different upsampling methods, but we believe these assessments should also be considered alongside quantitative evaluations in downstream tasks.

\subsection{Additional Efficiency Comparisons}

\input{figures_supp/scaling_full}

In Section \ref{sec:exppred}, we assessed the efficiency of UPLiFT in comparison to recent state-of-the-art cross-attention-based feature upsamplers \cite{couairon2025jafar, huang2025loftup, wimmer2025anyup}. We present additional efficiency results here. Along with reporting inference time, we also report the inference GPU memory usage, and we also present these results for two different GPU types: an NVIDIA A5000 with 24 GB of memory, and an NVIDIA H100 with 80 GB of memory.
In addition, we compare with AnyUp-v2 with its optional ``chunking mode'' enabled. This mode divides attention processing into smaller sub-chunks, which reduces maximum memory usage at the cost of higher inference times. We run this mode, denoted as ``AnyUp-v2-C'', with the recommend chunk size of $10$.

We plot these results in Figure \ref{fig:effsupp}. We observe similar trends in performance and scaling for both GPUs.
First, we see that LoftUp, JAFAR, and AnyUp-v1 show quadratic scaling in their inference time and memory usage.
Next, we see that UPLiFT and AnyUp-v2 both achieve linear scaling in inference time and memory. The improved efficiency of the NATTEN operations in AnyUp-v2 make it more competitive with UPLiFT, and while AnyUp-v2 can reach higher token counts, UPLiFT is still substantially faster in its functional range. We also see that chunking can further extend the max processing size for AnyUp-v2, though with a substantial increase in inference times.
Finally, we find that the inference optimizations of UPLIFT$_{\text{FAST}}$ greatly improve both its speed and memory scaling. For both GPUs, the maximum image processing size for UPLIFT$_{\text{FAST}}$ surpasses all other methods by a significant margin, while also maintaining fast speeds. On the H100, UPLIFT$_{\text{FAST}}$ can process images with resolutions over $3500{\times}3500$, and close to $50000$ visual tokens.

\subsection{Semantic Stability with UPLiFT}
\label{sec:suppsemdrift}
\input{figures_supp/semantic_drift_fig}

A critical issue with LiFT \cite{suri2024lift} is the occurrence of semantic drift through iterative feature upsampling. We present a visual comparison of semantic drift in LiFT compared with the semantic stability achieved by UPLiFT. Following \cite{fu2024featup}, we visualize the features of both methods using Principal Component Analysis (PCA) with 3 components. We extract the backbone features and all intermediate feature upsampling steps from both methods, and then perform a joint-PCA over the full collection. This means that consistent colors from one image to the next means consistent features too. We visualize the results in Figure \ref{fig:semdrift}.

Examining the PCA images for LiFT (row 1), we see that iterative upsampling leads to increasing degradation of the latents. The features become faded, murkier, and distorted through each step. Local image regions with consistent semantic content, like the dog's ear and nose region (row 3), do not maintain consistent and similar features. In comparison, UPLiFT (row 2) maintains consistent features through each upsampling step, without signs of internal distortions. In addition, the edges of objects are more distinctly defined than LiFT, and they match well to the original image, which enables better performance in tasks like semantic segmentation. Overall, these results demonstrate that our UPLiFT approach, through the use of Local Attender, is able to maintain consistent feature semantics through iterative upsampling steps to produce high-quality, pixel-dense features.

\subsection{Additional Visualization for Generative Tasks}
We provide further visualizations for UPLiFT applied to generative tasks.
Figure \ref{fig:super_res_qualitative} highlights UPLiFT's strong and efficient performance in $4{\times}$ super-resolution from $512{\times}512$ to $2048{\times}2048$. We find that UPLiFT adds only $8.47\%$ latency compared to bilinear upsampling in latent space, and yields far superior image quality to the naive upsampling method. 
Figure \ref{fig:diffusion_bulk_2048} includes $2048{\times}2048$ images, upscaled from Stable Diffusion 1.5, and Figure \ref{fig:diffusion_bulk_1024} shows the same at $1024{\times}1024$.
For the super-resolution task, we include samples from the evaluation datasets FacesHQ and LHQ using $2{\times}$ iterative upsampling twice, from $256{\times}256$ to $1024{\times}1024$ in Figure \ref{fig:super_res_faces_bulk} and Figure \ref{fig:lhq_bulk} respectively.

\input{figures_supp/super_res_qualitative}
\input{figures_supp/diffusion_samples_2048}
\input{figures_supp/diffusion_samples_1024}
\input{figures_supp/samples_faces}
\input{figures_supp/samples_lhq}

%% file: figures_supp/pca_comp.tex
\begin{figure*}[t]
    \centering
    \includegraphics[width=\linewidth]{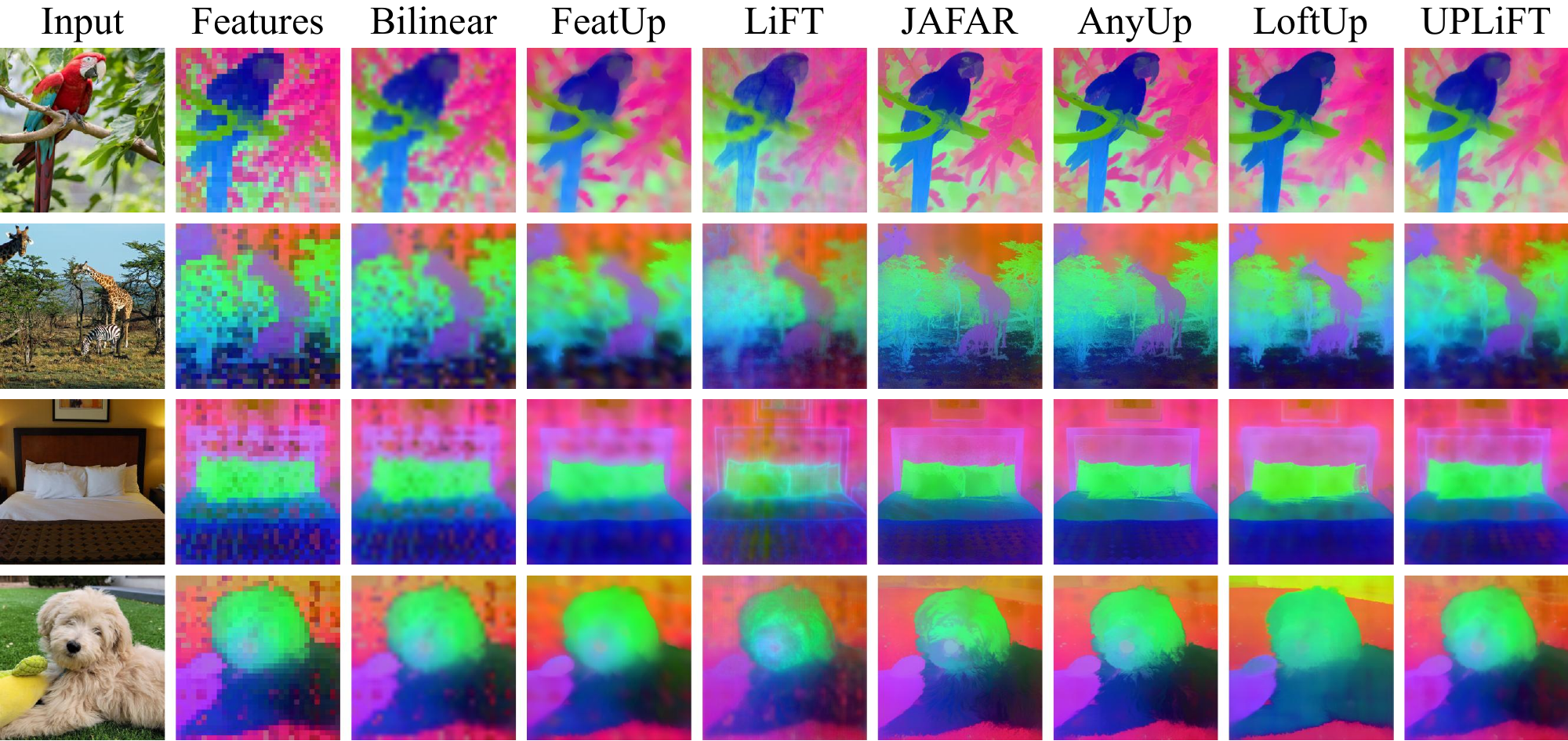}
    \vspace{-0.26in}
    \caption{\textbf{Qualitative Feature Comparison.} We present PCA visualizations for various upsampling methods applied to DINOv2 features.}
    \label{fig:pcacomp}
\end{figure*}

%% file: figures_supp/scaling_full.tex
    \begin{figure*}
        \vspace{0.7in}
        \centering
        \begin{subfigure}[b]{0.475\textwidth}
            \centering
            \includegraphics[width=\textwidth]{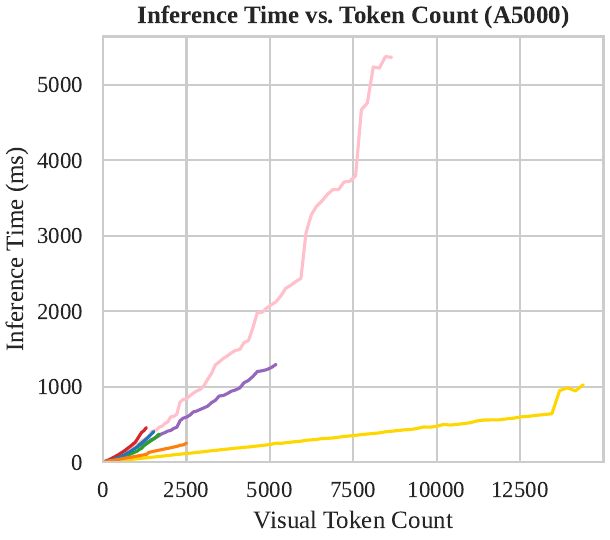}
        \end{subfigure}
        \hfill
        \begin{subfigure}[b]{0.475\textwidth}  
            \centering 
            \includegraphics[width=\textwidth]{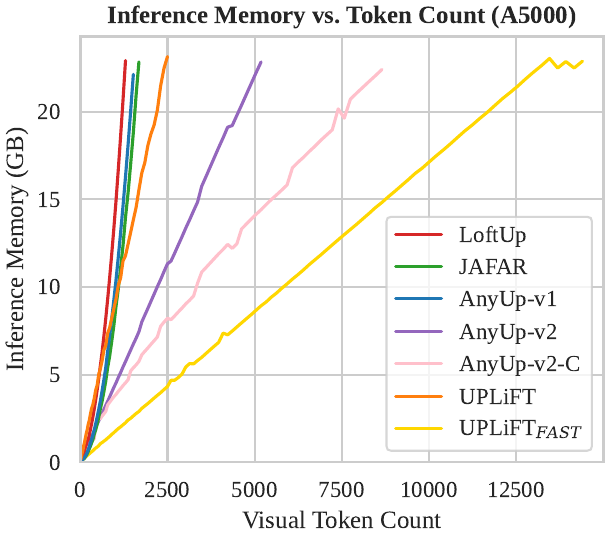}
        \end{subfigure}
        \vskip\baselineskip
        \begin{subfigure}[b]{0.475\textwidth}   
            \centering 
            \includegraphics[width=\textwidth]{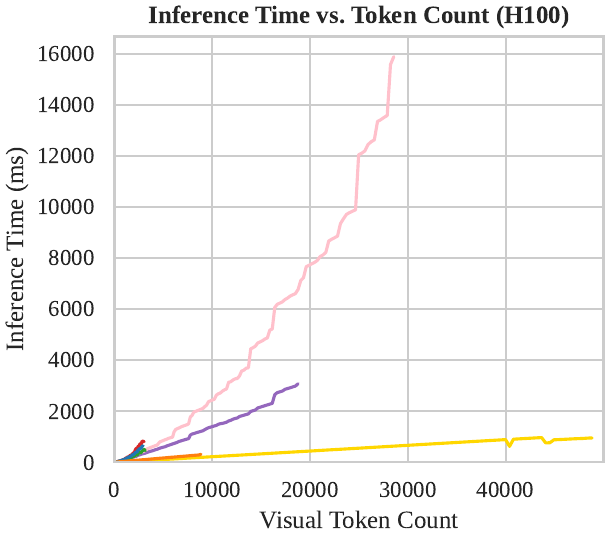}
        \end{subfigure}
        \hfill
        \begin{subfigure}[b]{0.475\textwidth}   
            \centering 
            \includegraphics[width=\textwidth]{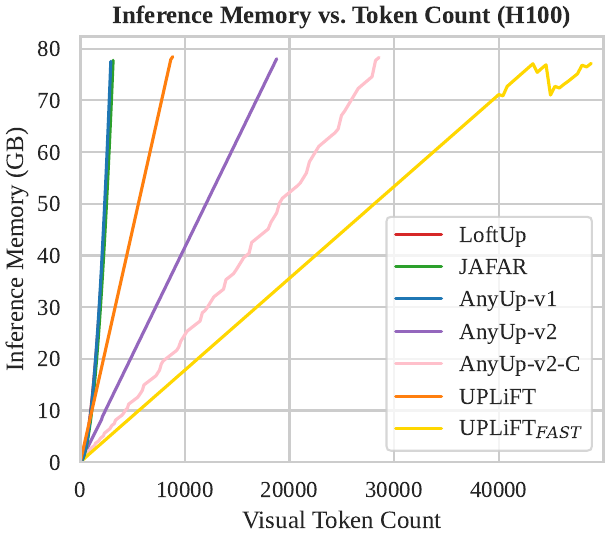}
        \end{subfigure}
        \caption{\textbf{Speed and Memory Usage Comparison.} We compare UPLiFT with recent cross-attention-based feature upsampling methods when running with gradually increasing image sizes. We report the average inference time and memory usage against the visual token count for two different GPU types: an NVIDIA A5000 with 24 GB of memory and an NVIDIA H100 with 80 GB of memory. Similar trends can be observed regardless of the GPU's max memory. UPLiFT maintains linear time and memory scaling with respect to the number of tokens, while the baseline methods LoftUp, JAFAR, and AnyUp-v1 show quadratic scaling. AnyUp-v2 achieves roughly linear scaling like UPLiFT, though UPLiFT still has faster speeds in the range it can operate in. UPLIFT$_{\text{FAST}}$ further accelerates UPLiFT inference and allows it to scale to even larger images. AnyUp's chunking mode also enables larger scaling, but at the cost of much slower inference speeds. The highest overall token counts are reached by UPLIFT$_{\text{FAST}}$ by a significant margin.}
        \label{fig:effsupp}
        \vspace{0.7in}
    \end{figure*}

%% file: figures_supp/semantic_drift_fig.tex
\begin{figure*}[t]
    \centering
    \vspace{0.35in}
    \includegraphics[width=1.0\linewidth]{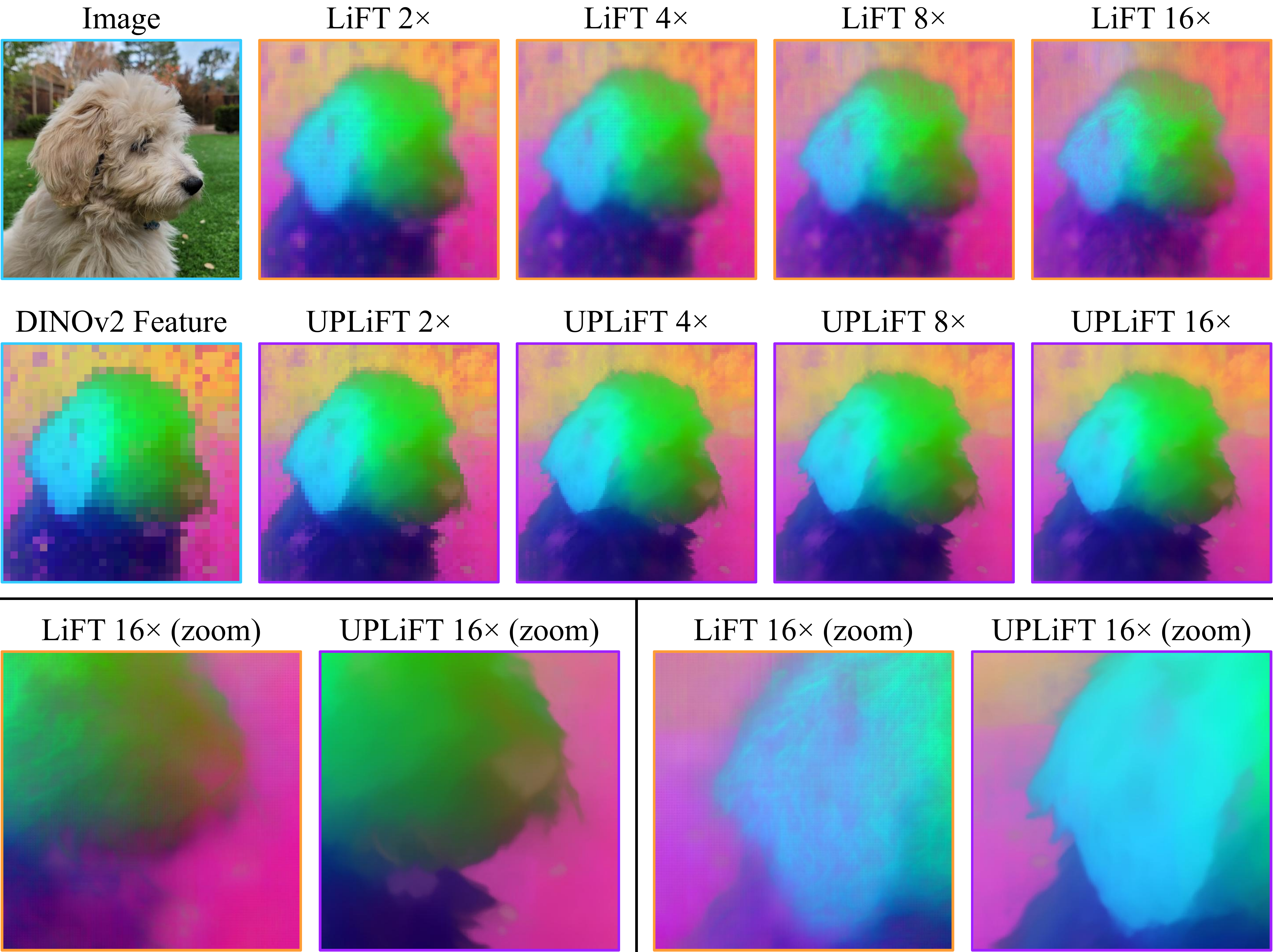}
    \caption{\textbf{Comparison of semantic drift in LiFT and semantic stability in UPLiFT.} We visualize intermediate feature upsampling steps through PCA, following \cite{fu2024featup}. LiFT shows signs of feature drift, with local features becoming murkier and more distorted in deeper steps. This drift can lead to poor performance in downstream tasks, as the strength of the original backbone representation is lost. UPLiFT maintains consistent feature semantics thanks to our Local Attender, and local object regions maintain consistent features (coloration) across all upsampling stages.}
    \label{fig:semdrift}
\end{figure*}

%% file: figures_supp/super_res_qualitative.tex
\begin{figure*}[p]
    \centering
    \includegraphics[width=0.9\linewidth]{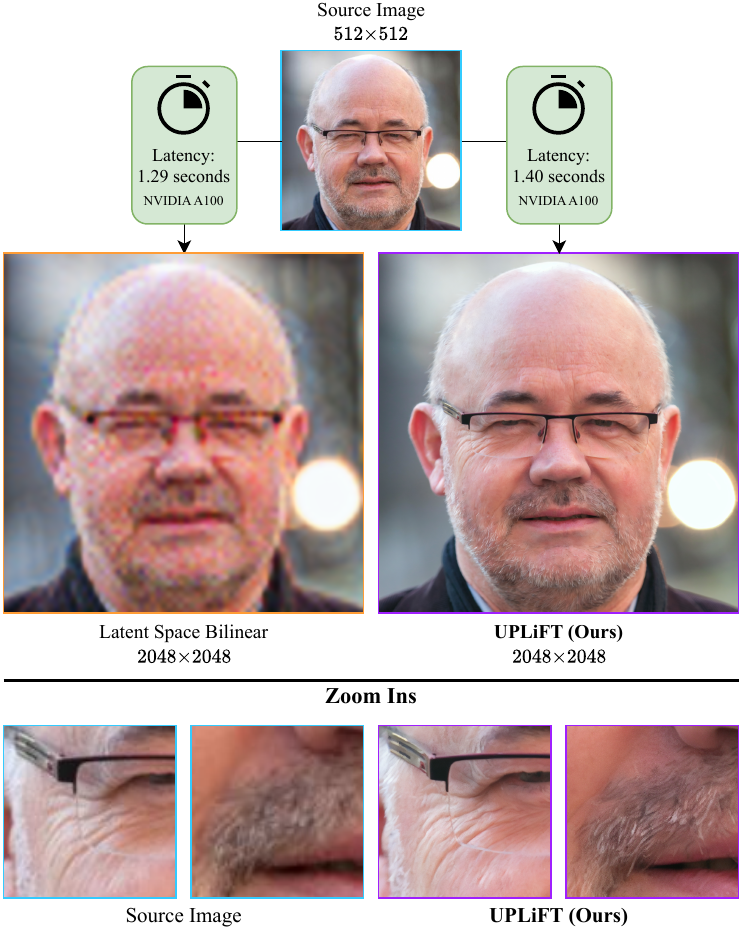}
    \caption{\textbf{UPLiFT \vs Latent Space Bilinear $4{\times}$ upsampling for image super-resolution.}
    We compare the latency of UPLiFT versus applying bilinear upsampling in latent space for image super-resolution to demonstrate UPLiFT's state-of-the-art efficiency.
    While only $8.47\%$ slower end-to-end, UPLiFT produces significantly better visual fidelity, as shown by the zoomed-in views (bottom) which display the source low-resolution image compared with UPLiFT's super-resolution image.
    The low-resolution image is selected from the FacesHQ dataset and bilinearly downscaled from $1024{\times}1024$ to $512{\times}512$ before UPLiFT is applied.
    Best viewed zoomed in.
    }
    \label{fig:super_res_qualitative}
\end{figure*}

%% file: figures_supp/diffusion_samples_2048.tex
\begin{figure*}[p]
    \centering
    \includegraphics[width=\linewidth]{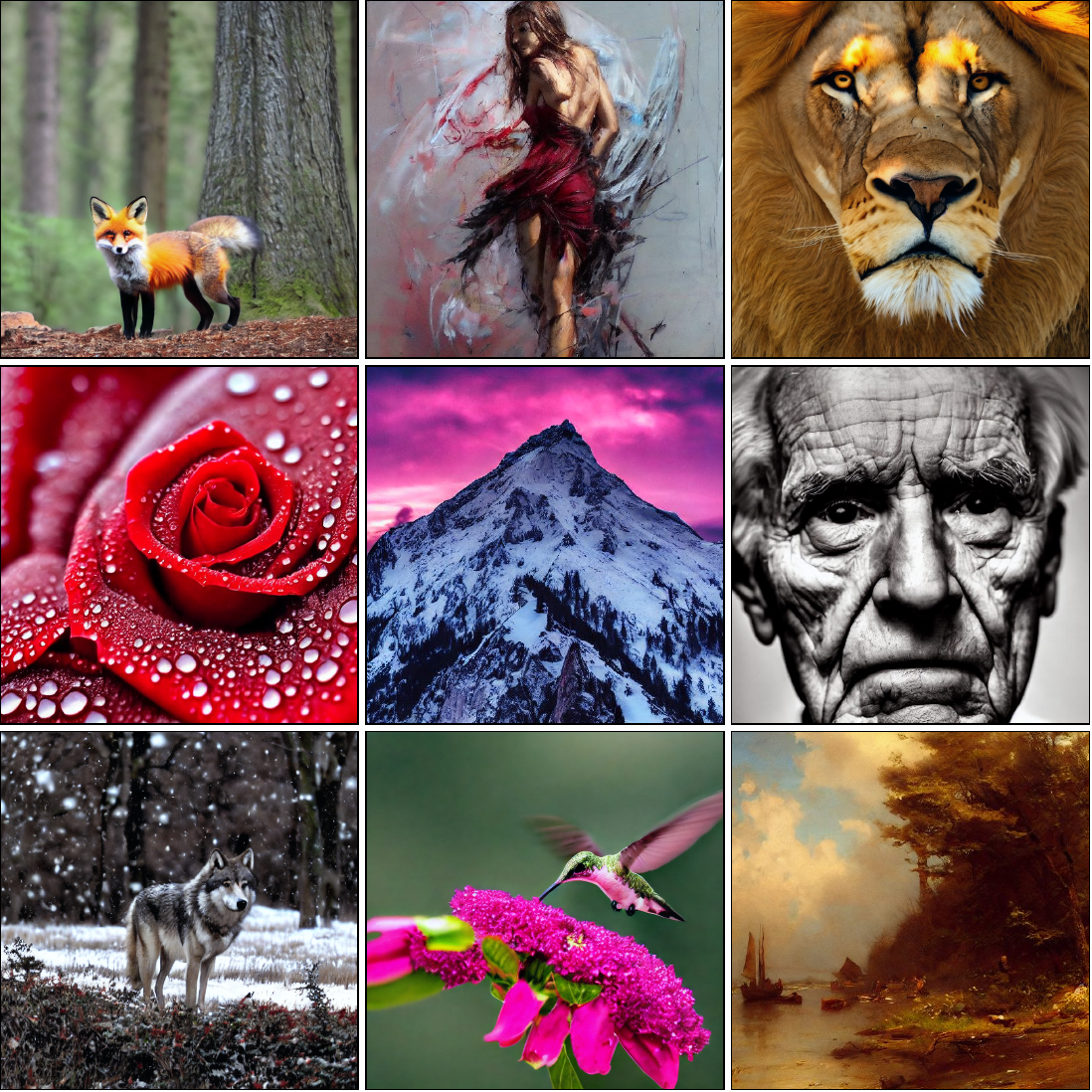}
    \caption{\textbf{UPLiFT $\mathbf{512{\times}512 \to 2048{\times}2048}$ upsampled images using Stable Diffusion 1.5.}
    We apply our VAE UPLiFT model to this task in a $4{\times}$ upsampling configuration.
    UPLiFT upsamples latents corresponding to $512{\times}512$ images generated using 50 diffusion steps on Stable Diffusion 1.5. 
    Best viewed zoomed in.
    }
    \label{fig:diffusion_bulk_2048}
\end{figure*}

%% file: figures_supp/diffusion_samples_1024.tex
\begin{figure*}[p]
    \centering
    \includegraphics[width=\linewidth]{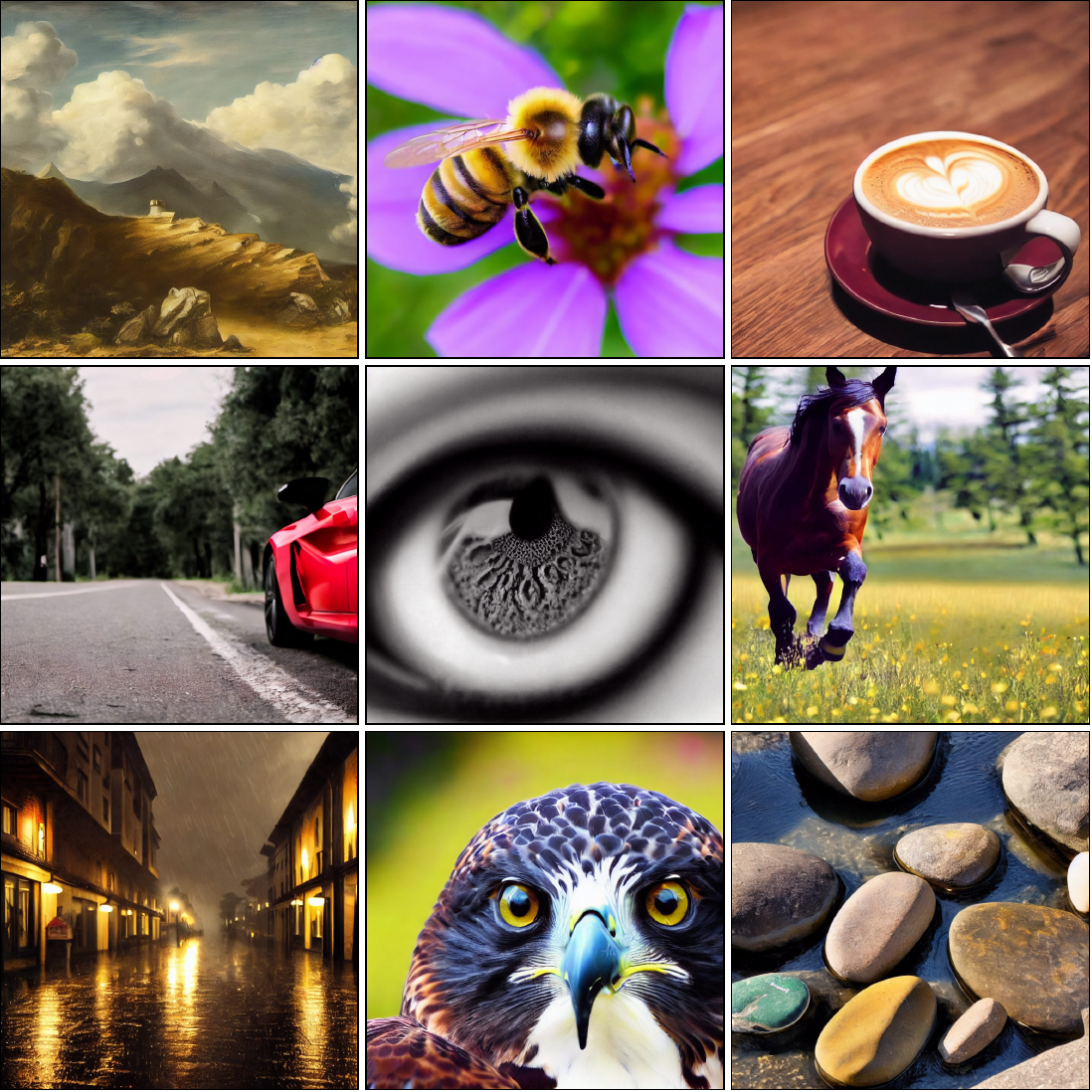}
    \caption{\textbf{UPLiFT $\mathbf{512{\times}512 \to 1024{\times}1024}$ upsampled images using Stable Diffusion 1.5.}
    We apply our VAE UPLiFT model to this task in a $2{\times}$ upsampling configuration.
    UPLiFT upsamples latents corresponding to $512{\times}512$ images generated using 50 diffusion steps on Stable Diffusion 1.5, and the end-to-end latency is 2.75 seconds on an NVIDIA A100 GPU. The UPLiFT model itself takes only 104 milliseconds of this time. 
    Best viewed zoomed in.
    }
    \label{fig:diffusion_bulk_1024}
\end{figure*}

%% file: figures_supp/samples_faces.tex
\begin{figure*}[p]
    \centering
    \includegraphics[width=1.0\linewidth]{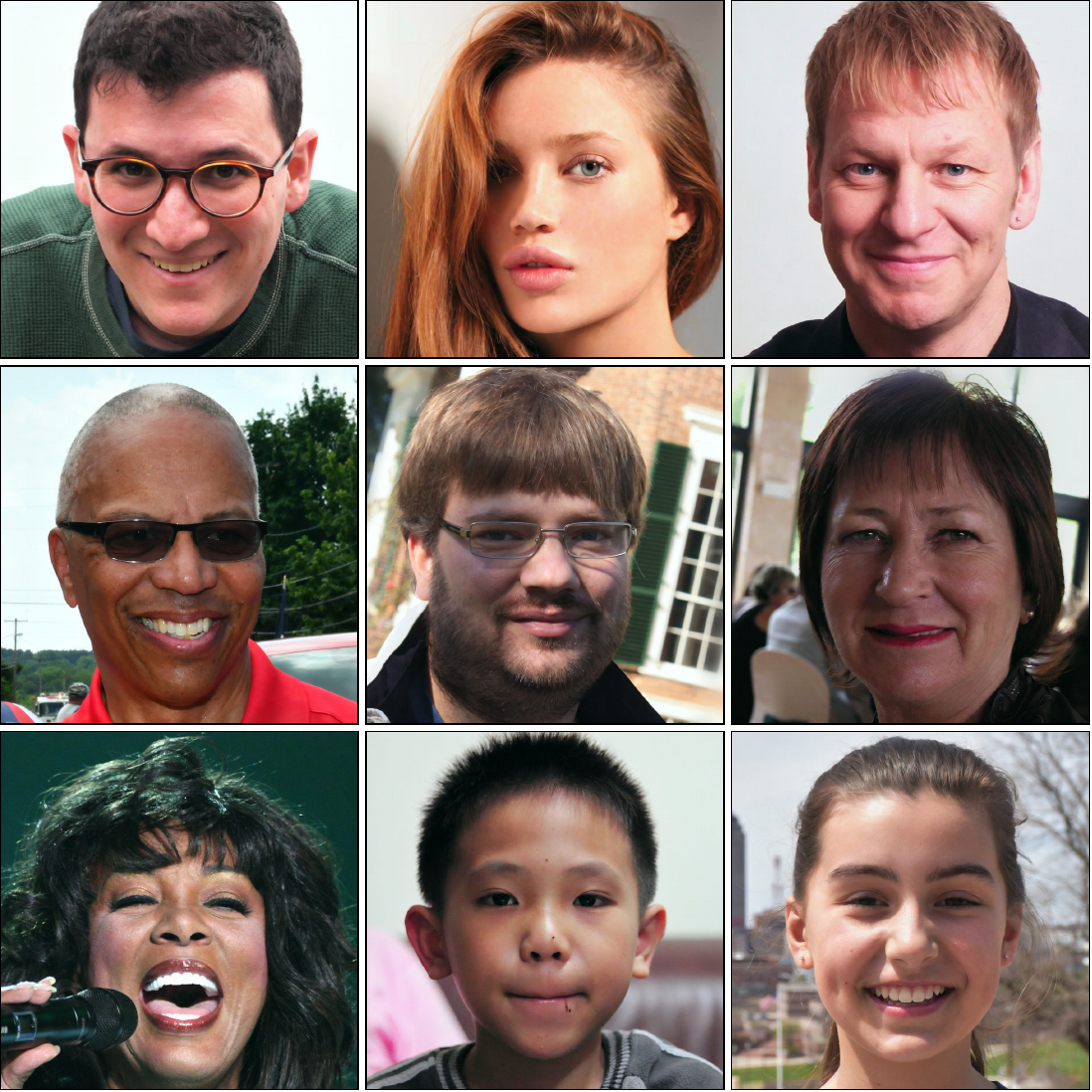}
    \caption{\textbf{UPLiFT $\mathbf{256{\times}256 \to 1024{\times}1024}$ super-resolution samples from FacesHQ.}
    We use our VAE UPLiFT model that is \emph{not fine-tuned} for image super-resolution and is only trained in latent space.
    Our end-to-end upsampling time is only \textbf{270.9 milliseconds} on an NVIDIA A100 GPU.
    Best viewed zoomed in.
    }
    \label{fig:super_res_faces_bulk}
\end{figure*}

%% file: figures_supp/samples_lhq.tex
\begin{figure*}[p]
    \centering
    \includegraphics[width=1.0\linewidth]{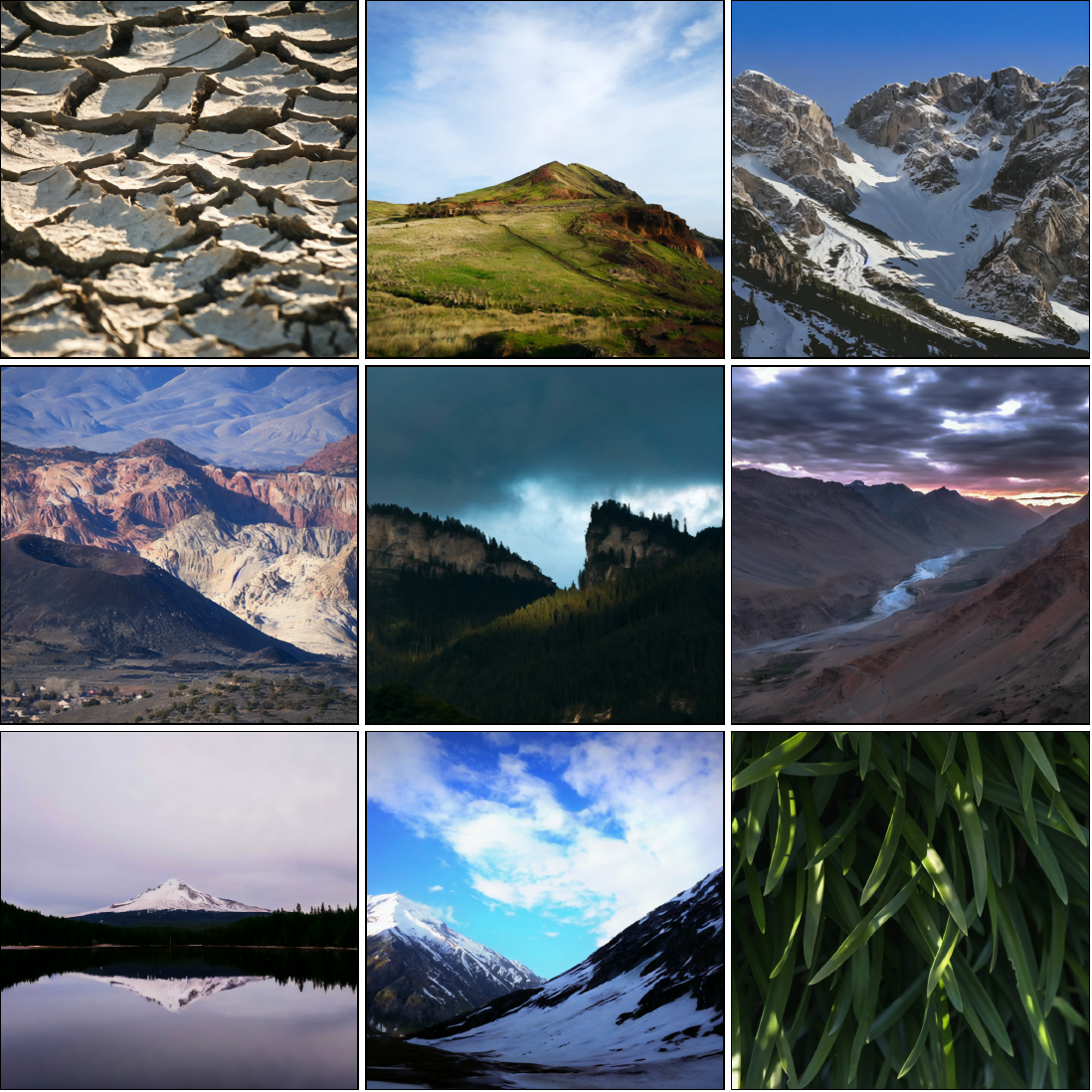}
    \caption{\textbf{UPLiFT $\mathbf{256{\times}256 \to 1024{\times}1024}$ super-resolution samples from LHQ.}
    We use our VAE UPLiFT model, which is trained as a generalist model and is not specifically fine-tuned for this dataset.
    The LHQ dataset presents a greater challenge than FacesHQ, based on the diversity of visual textures present. Despite this challenge, we see good performance with our generalist UPLiFT.
    In comparison, \cite{schusterbauer2024fmboost} uses a fine-tuned model with $3{\times}$ the parameter count for evaluations on LHQ versus FacesHQ.
    Best viewed zoomed in.
    }
    \label{fig:lhq_bulk}
\end{figure*}

%% file: main.bib
@String(CVPR= {IEEE Conf. Comput. Vis. Pattern Recog.})

@String(TOG= {ACM Trans. Graph.})

@String(CVPR  = {CVPR})

@String(TOG   = {ACM TOG})

@inproceedings{suri2024lift,
  title={Lift: A surprisingly simple lightweight feature transform for dense vit descriptors},
  author={Suri, Saksham and Walmer, Matthew and Gupta, Kamal and Shrivastava, Abhinav},
  booktitle={European Conference on Computer Vision},
  pages={110--128},
  year={2024},
  organization={Springer}
}

@article{fu2024featup,
  title={Featup: A model-agnostic framework for features at any resolution},
  author={Fu, Stephanie and Hamilton, Mark and Brandt, Laura and Feldman, Axel and Zhang, Zhoutong and Freeman, William T},
  journal={arXiv preprint arXiv:2403.10516},
  year={2024}
}

@article{couairon2025jafar,
  title={JAFAR: Jack up Any Feature at Any Resolution},
  author={Couairon, Paul and Chambon, Loick and Serrano, Louis and Haugeard, Jean-Emmanuel and Cord, Matthieu and Thome, Nicolas},
  journal={arXiv preprint arXiv:2506.11136},
  year={2025}
}

@article{huang2025loftup,
  title={LoftUp: Learning a Coordinate-Based Feature Upsampler for Vision Foundation Models},
  author={Huang, Haiwen and Chen, Anpei and Havrylov, Volodymyr and Geiger, Andreas and Zhang, Dan},
  journal={arXiv preprint arXiv:2504.14032},
  year={2025}
}

@article{wimmer2025anyup,
  title={Anyup: Universal feature upsampling},
  author={Wimmer, Thomas and Truong, Prune and Rakotosaona, Marie-Julie and Oechsle, Michael and Tombari, Federico and Schiele, Bernt and Lenssen, Jan Eric},
  journal={arXiv preprint arXiv:2510.12764},
  year={2025}
}

@inproceedings{schusterbauer2024fmboost,
  title={Fmboost: Boosting latent diffusion with flow matching},
  author={Schusterbauer, Johannes and Gui, Ming and Ma, Pingchuan and Stracke, Nick and Baumann, Stefan Andreas and Hu, Vincent Tao and Ommer, Bj{\"o}rn},
  booktitle={European Conference on Computer Vision},
  pages={338--355},
  year={2024},
  organization={Springer}
}

@inproceedings{lin2014microsoft,
  title={Microsoft coco: Common objects in context},
  author={Lin, Tsung-Yi and Maire, Michael and Belongie, Serge and Hays, James and Perona, Pietro and Ramanan, Deva and Doll{\'a}r, Piotr and Zitnick, C Lawrence},
  booktitle={European conference on computer vision},
  pages={740--755},
  year={2014},
  organization={Springer}
}

@article{everingham2015pascal,
  title={The pascal visual object classes challenge: A retrospective},
  author={Everingham, Mark and Eslami, SM Ali and Van Gool, Luc and Williams, Christopher KI and Winn, John and Zisserman, Andrew},
  journal={International journal of computer vision},
  volume={111},
  number={1},
  pages={98--136},
  year={2015},
  publisher={Springer}
}

@article{zhou2019semantic,
  title={Semantic understanding of scenes through the ade20k dataset},
  author={Zhou, Bolei and Zhao, Hang and Puig, Xavier and Xiao, Tete and Fidler, Sanja and Barriuso, Adela and Torralba, Antonio},
  journal={International Journal of Computer Vision},
  volume={127},
  number={3},
  pages={302--321},
  year={2019},
  publisher={Springer}
}

@inproceedings{cordts2016cityscapes,
  title={The cityscapes dataset for semantic urban scene understanding},
  author={Cordts, Marius and Omran, Mohamed and Ramos, Sebastian and Rehfeld, Timo and Enzweiler, Markus and Benenson, Rodrigo and Franke, Uwe and Roth, Stefan and Schiele, Bernt},
  booktitle={Proceedings of the IEEE conference on computer vision and pattern recognition},
  pages={3213--3223},
  year={2016}
}

@article{saharia2022image,
  title={Image super-resolution via iterative refinement},
  author={Saharia, Chitwan and Ho, Jonathan and Chan, William and Salimans, Tim and Fleet, David J and Norouzi, Mohammad},
  journal={IEEE transactions on pattern analysis and machine intelligence},
  volume={45},
  number={4},
  pages={4713--4726},
  year={2022},
  publisher={IEEE}
}

@inproceedings{walmer2023teaching,
  title={Teaching matters: Investigating the role of supervision in vision transformers},
  author={Walmer, Matthew and Suri, Saksham and Gupta, Kamal and Shrivastava, Abhinav},
  booktitle={Proceedings of the IEEE/CVF Conference on Computer Vision and Pattern Recognition},
  pages={7486--7496},
  year={2023}
}

@inproceedings{caron2021emerging,
  title={Emerging properties in self-supervised vision transformers},
  author={Caron, Mathilde and Touvron, Hugo and Misra, Ishan and J{\'e}gou, Herv{\'e} and Mairal, Julien and Bojanowski, Piotr and Joulin, Armand},
  booktitle={Proceedings of the IEEE/CVF international conference on computer vision},
  pages={9650--9660},
  year={2021}
}

@article{oquab2023dinov2,
  title={Dinov2: Learning robust visual features without supervision},
  author={Oquab, Maxime and Darcet, Timoth{\'e}e and Moutakanni, Th{\'e}o and Vo, Huy and Szafraniec, Marc and Khalidov, Vasil and Fernandez, Pierre and Haziza, Daniel and Massa, Francisco and El-Nouby, Alaaeldin and others},
  journal={arXiv preprint arXiv:2304.07193},
  year={2023}
}

@article{simeoni2025dinov3,
  title={Dinov3},
  author={Sim{\'e}oni, Oriane and Vo, Huy V and Seitzer, Maximilian and Baldassarre, Federico and Oquab, Maxime and Jose, Cijo and Khalidov, Vasil and Szafraniec, Marc and Yi, Seungeun and Ramamonjisoa, Micha{\"e}l and others},
  journal={arXiv preprint arXiv:2508.10104},
  year={2025}
}

@article{dosovitskiy2020image,
  title={An image is worth 16x16 words: Transformers for image recognition at scale},
  author={Dosovitskiy, Alexey},
  journal={arXiv preprint arXiv:2010.11929},
  year={2020}
}

@article{amir2021deep,
  title={Deep vit features as dense visual descriptors},
  author={Amir, Shir and Gandelsman, Yossi and Bagon, Shai and Dekel, Tali},
  journal={arXiv preprint arXiv:2112.05814},
  volume={2},
  number={3},
  pages={4},
  year={2021}
}

@article{kingma2013auto,
  title={Auto-encoding variational bayes},
  author={Kingma, Diederik P and Welling, Max},
  journal={arXiv preprint arXiv:1312.6114},
  year={2013}
}

@article{chen2020improved,
  title={Improved baselines with momentum contrastive learning},
  author={Chen, Xinlei and Fan, Haoqi and Girshick, Ross and He, Kaiming},
  journal={arXiv preprint arXiv:2003.04297},
  year={2020}
}

@inproceedings{chen2021empirical,
  title={An empirical study of training self-supervised vision transformers},
  author={Chen, Xinlei and Xie, Saining and He, Kaiming},
  booktitle={Proceedings of the IEEE/CVF international conference on computer vision},
  pages={9640--9649},
  year={2021}
}

@inproceedings{he2020momentum,
  title={Momentum contrast for unsupervised visual representation learning},
  author={He, Kaiming and Fan, Haoqi and Wu, Yuxin and Xie, Saining and Girshick, Ross},
  booktitle={Proceedings of the IEEE/CVF conference on computer vision and pattern recognition},
  pages={9729--9738},
  year={2020}
}

@inproceedings{he2022masked,
  title={Masked autoencoders are scalable vision learners},
  author={He, Kaiming and Chen, Xinlei and Xie, Saining and Li, Yanghao and Doll{\'a}r, Piotr and Girshick, Ross},
  booktitle={Proceedings of the IEEE/CVF conference on computer vision and pattern recognition},
  pages={16000--16009},
  year={2022}
}

@inproceedings{radford2021learning,
  title={Learning transferable visual models from natural language supervision},
  author={Radford, Alec and Kim, Jong Wook and Hallacy, Chris and Ramesh, Aditya and Goh, Gabriel and Agarwal, Sandhini and Sastry, Girish and Askell, Amanda and Mishkin, Pamela and Clark, Jack and others},
  booktitle={International conference on machine learning},
  pages={8748--8763},
  year={2021},
  organization={PmLR}
}

@inproceedings{zhai2023sigmoid,
  title={Sigmoid loss for language image pre-training},
  author={Zhai, Xiaohua and Mustafa, Basil and Kolesnikov, Alexander and Beyer, Lucas},
  booktitle={Proceedings of the IEEE/CVF international conference on computer vision},
  pages={11975--11986},
  year={2023}
}

@article{tschannen2025siglip,
  title={Siglip 2: Multilingual vision-language encoders with improved semantic understanding, localization, and dense features},
  author={Tschannen, Michael and Gritsenko, Alexey and Wang, Xiao and Naeem, Muhammad Ferjad and Alabdulmohsin, Ibrahim and Parthasarathy, Nikhil and Evans, Talfan and Beyer, Lucas and Xia, Ye and Mustafa, Basil and others},
  journal={arXiv preprint arXiv:2502.14786},
  year={2025}
}

@article{kopf2007joint,
  title={Joint bilateral upsampling},
  author={Kopf, Johannes and Cohen, Michael F and Lischinski, Dani and Uyttendaele, Matt},
  journal={ACM Transactions on Graphics (ToG)},
  volume={26},
  number={3},
  pages={96--es},
  year={2007},
  publisher={ACM New York, NY, USA}
}

@inproceedings{chen2021learning,
  title={Learning continuous image representation with local implicit image function},
  author={Chen, Yinbo and Liu, Sifei and Wang, Xiaolong},
  booktitle={Proceedings of the IEEE/CVF conference on computer vision and pattern recognition},
  pages={8628--8638},
  year={2021}
}

@article{vaswani2017attention,
  title={Attention is all you need},
  author={Vaswani, Ashish and Shazeer, Noam and Parmar, Niki and Uszkoreit, Jakob and Jones, Llion and Gomez, Aidan N and Kaiser, {\L}ukasz and Polosukhin, Illia},
  journal={Advances in neural information processing systems},
  volume={30},
  year={2017}
}

@inproceedings{rombach2022high,
  title={High-resolution image synthesis with latent diffusion models},
  author={Rombach, Robin and Blattmann, Andreas and Lorenz, Dominik and Esser, Patrick and Ommer, Bj{\"o}rn},
  booktitle={Proceedings of the IEEE/CVF conference on computer vision and pattern recognition},
  pages={10684--10695},
  year={2022}
}

@inproceedings{chen2018fsrnet,
  title={Fsrnet: End-to-end learning face super-resolution with facial priors},
  author={Chen, Yu and Tai, Ying and Liu, Xiaoming and Shen, Chunhua and Yang, Jian},
  booktitle={Proceedings of the IEEE conference on computer vision and pattern recognition},
  pages={2492--2501},
  year={2018}
}

@inproceedings{dahl2017pixel,
  title={Pixel recursive super resolution},
  author={Dahl, Ryan and Norouzi, Mohammad and Shlens, Jonathon},
  booktitle={Proceedings of the IEEE international conference on computer vision},
  pages={5439--5448},
  year={2017}
}

@inproceedings{ledig2017photo,
  title={Photo-realistic single image super-resolution using a generative adversarial network},
  author={Ledig, Christian and Theis, Lucas and Husz{\'a}r, Ferenc and Caballero, Jose and Cunningham, Andrew and Acosta, Alejandro and Aitken, Andrew and Tejani, Alykhan and Totz, Johannes and Wang, Zehan and others},
  booktitle={Proceedings of the IEEE conference on computer vision and pattern recognition},
  pages={4681--4690},
  year={2017}
}

@inproceedings{menon2020pulse,
  title={Pulse: Self-supervised photo upsampling via latent space exploration of generative models},
  author={Menon, Sachit and Damian, Alexandru and Hu, Shijia and Ravi, Nikhil and Rudin, Cynthia},
  booktitle={Proceedings of the ieee/cvf conference on computer vision and pattern recognition},
  pages={2437--2445},
  year={2020}
}

@article{hwang2024upsample,
  title={Upsample guidance: Scale up diffusion models without training},
  author={Hwang, Juno and Park, Yong-Hyun and Jo, Junghyo},
  journal={arXiv preprint arXiv:2404.01709},
  year={2024}
}

@inproceedings{qiu2025freescale,
  title={Freescale: Unleashing the resolution of diffusion models via tuning-free scale fusion},
  author={Qiu, Haonan and Zhang, Shiwei and Wei, Yujie and Chu, Ruihang and Yuan, Hangjie and Wang, Xiang and Zhang, Yingya and Liu, Ziwei},
  booktitle={Proceedings of the IEEE/CVF International Conference on Computer Vision},
  pages={16893--16903},
  year={2025}
}

@article{yang2025rectifiedhr,
  title={Rectifiedhr: Enable efficient high-resolution image generation via energy rectification},
  author={Yang, Zhen and Shen, Guibao and Hou, Liang and Liu, Mushui and Wang, Luozhou and Tao, Xin and Wan, Pengfei and Zhang, Di and Chen, Ying-Cong},
  journal={arXiv e-prints},
  pages={arXiv--2503},
  year={2025}
}

@article{tragakis2024one,
  title={Is one gpu enough? pushing image generation at higher-resolutions with foundation models},
  author={Tragakis, Athanasios and Aversa, Marco and Kaul, Chaitanya and Murray-Smith, Roderick and Faccio, Daniele},
  journal={arXiv preprint arXiv:2406.07251},
  volume={2},
  number={3},
  pages={5},
  year={2024}
}

@article{lin2025accdiffusion,
  title={AccDiffusion v2: Towards More Accurate Higher-Resolution Diffusion Extrapolation},
  author={Lin, Zhihang and Lin, Mingbao and Zhan, Wengyi and Ji, Rongrong},
  journal={IEEE Transactions on Pattern Analysis and Machine Intelligence},
  year={2025},
  publisher={IEEE}
}

@article{li2024asgdiffusion,
  title={ASGDiffusion: Parallel High-Resolution Generation with Asynchronous Structure Guidance},
  author={Li, Yuming and Jia, Peidong and Hong, Daiwei and Jia, Yueru and She, Qi and Zhao, Rui and Lu, Ming and Zhang, Shanghang},
  journal={arXiv preprint arXiv:2412.06163},
  year={2024}
}

@misc{labs2025flux1kontextflowmatching,
      title={FLUX.1 Kontext: Flow Matching for In-Context Image Generation and Editing in Latent Space},
      author={Black Forest Labs and Stephen Batifol and Andreas Blattmann and Frederic Boesel and Saksham Consul and Cyril Diagne and Tim Dockhorn and Jack English and Zion English and Patrick Esser and Sumith Kulal and Kyle Lacey and Yam Levi and Cheng Li and Dominik Lorenz and Jonas Müller and Dustin Podell and Robin Rombach and Harry Saini and Axel Sauer and Luke Smith},
      year={2025},
      eprint={2506.15742},
      archivePrefix={arXiv},
      primaryClass={cs.GR},
      url={https://arxiv.org/abs/2506.15742},
}

@article{albergo2023stochastic,
  title={Stochastic interpolants: A unifying framework for flows and diffusions},
  author={Albergo, Michael S and Boffi, Nicholas M and Vanden-Eijnden, Eric},
  journal={arXiv preprint arXiv:2303.08797},
  year={2023}
}

@article{lipman2022flow,
  title={Flow matching for generative modeling},
  author={Lipman, Yaron and Chen, Ricky TQ and Ben-Hamu, Heli and Nickel, Maximilian and Le, Matt},
  journal={arXiv preprint arXiv:2210.02747},
  year={2022}
}

@article{liu2022flow,
  title={Flow straight and fast: Learning to generate and transfer data with rectified flow},
  author={Liu, Xingchao and Gong, Chengyue and Liu, Qiang},
  journal={arXiv preprint arXiv:2209.03003},
  year={2022}
}

@misc{rombach2021highresolution,
      title={High-Resolution Image Synthesis with Latent Diffusion Models}, 
      author={Robin Rombach and Andreas Blattmann and Dominik Lorenz and Patrick Esser and Björn Ommer},
      year={2021},
      eprint={2112.10752},
      archivePrefix={arXiv},
      primaryClass={cs.CV}
}

@inproceedings{wang2019carafe,
  title={Carafe: Content-aware reassembly of features},
  author={Wang, Jiaqi and Chen, Kai and Xu, Rui and Liu, Ziwei and Loy, Chen Change and Lin, Dahua},
  booktitle={Proceedings of the IEEE/CVF international conference on computer vision},
  pages={3007--3016},
  year={2019}
}

@article{lu2022sapa,
  title={SAPA: Similarity-aware point affiliation for feature upsampling},
  author={Lu, Hao and Liu, Wenze and Ye, Zixuan and Fu, Hongtao and Liu, Yuliang and Cao, Zhiguo},
  journal={Advances in Neural Information Processing Systems},
  volume={35},
  pages={20889--20901},
  year={2022}
}

@inproceedings{liu2023learning,
  title={Learning to upsample by learning to sample},
  author={Liu, Wenze and Lu, Hao and Fu, Hongtao and Cao, Zhiguo},
  booktitle={Proceedings of the IEEE/CVF international conference on computer vision},
  pages={6027--6037},
  year={2023}
}

@article{zhou2024refreshed,
  title={A refreshed similarity-based upsampler for direct high-ratio feature upsampling},
  author={Zhou, Minghao and Wang, Hong and Zheng, Yefeng and Meng, Deyu},
  journal={arXiv preprint arXiv:2407.02283},
  year={2024}
}

@inproceedings{deng2009imagenet,
  title={Imagenet: A large-scale hierarchical image database},
  author={Deng, Jia and Dong, Wei and Socher, Richard and Li, Li-Jia and Li, Kai and Fei-Fei, Li},
  booktitle={2009 IEEE conference on computer vision and pattern recognition},
  pages={248--255},
  year={2009},
  organization={Ieee}
}

@misc{unsplash_data,
  author       = {{Unsplash}},
  title        = {{Unsplash Full, Lite Dataset} 1.3.0},
  year         = {2025},
  url          = {https://unsplash.com/data},
  note         = {Accessed: 14 November 2025}
}

@article{wang2004image,
  title={Image quality assessment: from error visibility to structural similarity},
  author={Wang, Zhou and Bovik, Alan C and Sheikh, Hamid R and Simoncelli, Eero P},
  journal={IEEE transactions on image processing},
  volume={13},
  number={4},
  pages={600--612},
  year={2004},
  publisher={IEEE}
}

@article{ba2016layer,
  title={Layer normalization},
  author={Ba, Jimmy Lei and Kiros, Jamie Ryan and Hinton, Geoffrey E},
  journal={arXiv preprint arXiv:1607.06450},
  year={2016}
}

@inproceedings{ioffe2015batch,
  title={Batch normalization: Accelerating deep network training by reducing internal covariate shift},
  author={Ioffe, Sergey and Szegedy, Christian},
  booktitle={International conference on machine learning},
  pages={448--456},
  year={2015},
  organization={pmlr}
}

@article{podell2023sdxl,
  title={Sdxl: Improving latent diffusion models for high-resolution image synthesis},
  author={Podell, Dustin and English, Zion and Lacey, Kyle and Blattmann, Andreas and Dockhorn, Tim and M{\"u}ller, Jonas and Penna, Joe and Rombach, Robin},
  journal={arXiv preprint arXiv:2307.01952},
  year={2023}
}

@article{luo2023lcm,
  title={Lcm-lora: A universal stable-diffusion acceleration module},
  author={Luo, Simian and Tan, Yiqin and Patil, Suraj and Gu, Daniel and Von Platen, Patrick and Passos, Apolin{\'a}rio and Huang, Longbo and Li, Jian and Zhao, Hang},
  journal={arXiv preprint arXiv:2311.05556},
  year={2023}
}

@inproceedings{hassani2023neighborhood,
  title        = {Neighborhood Attention Transformer},
  author       = {Ali Hassani and Steven Walton and Jiachen Li and Shen Li and Humphrey Shi},
  year         = 2023,
  booktitle    = {IEEE/CVF Conference on Computer Vision and Pattern Recognition (CVPR)}
}

@inproceedings{kirillov2023segment,
  title={Segment anything},
  author={Kirillov, Alexander and Mintun, Eric and Ravi, Nikhila and Mao, Hanzi and Rolland, Chloe and Gustafson, Laura and Xiao, Tete and Whitehead, Spencer and Berg, Alexander C and Lo, Wan-Yen and others},
  booktitle={Proceedings of the IEEE/CVF international conference on computer vision},
  pages={4015--4026},
  year={2023}
}
